\documentclass[final,5p,times,authoryear]{elsarticle}

\usepackage[T1]{fontenc}
\usepackage[utf8]{inputenc}
\usepackage{graphicx}
\usepackage{amsmath}
\usepackage{amssymb}
\usepackage{booktabs}
\usepackage{longtable}
\usepackage{array}
\usepackage{multirow}
\usepackage{url}
\usepackage{microtype}
\newcolumntype{L}[1]{>{\raggedright\arraybackslash}p{#1}}
\newcolumntype{C}[1]{>{\centering\arraybackslash}p{#1}}
\newcolumntype{R}[1]{>{\raggedleft\arraybackslash}p{#1}}

\journal{arXiv}

\begin{document}

\begin{frontmatter}

\title{Listening to the Workforce: Measuring Construction Worker Safety Attitudes from Social Media Discourse Using LLMs}

\author[tamu]{Farouq Sammour}
\author[tamu]{Yuxin Zhang}
\author[tamu]{Zhenyu Zhang\corref{cor1}}
\ead{z.zhang@tamu.edu}
\cortext[cor1]{Corresponding author.}
\affiliation[tamu]{organization={Department of Construction Science, College of Architecture, Texas A\&M University},
  city={College Station},
  country={USA}}

\begin{abstract}
Worker safety attitudes are key determinants of whether protective practices are applied or bypassed on construction sites. Yet measuring them at scale has remained out of reach. Safety attitudes are multidimensional, vary across topics, and surface most candidly in workers' own conversations. This study created and validated the Construction Safety Attitude Framework (CSAF), which integrates two components: a theory-grounded structure that characterizes safety attitudes along eight dimensions, and an operational codebook for measuring them in worker naturalistic discourse. Applying CSAF to 250 posts and comments from the r/Construction community on Reddit, trained coders reached strong agreement (Krippendorff's $\alpha = 0.85$). Pairwise lift and conditional probability confirmed that the eight dimensions are related yet distinct. To apply the framework across large volumes of discourse, CSAF was operationalized through a large language model (LLM) classifier. On 450 r/Construction contributions, the classifier reproduced expert human coding (Cohen's $\kappa = 0.90$, precision $= 0.98$, recall $= 0.98$), and on 400 contributions from r/Roofing it retained that accuracy after transfer to a different trade community ($\kappa = 0.89$, precision $= 0.98$, recall $= 0.97$). A proof-of-value case study then applied the validated classifier to 10,346 contributions from r/Roofing, demonstrating that CSAF can distinguish multidimensional attitudes by safety topic, track how they shift over time, and trace the reasoning behind unfavorable ones. The study therefore provides a theoretically grounded, empirically vetted instrument for examining safety attitudes, offering a basis for targeted interventions that address the attitudes underlying unsafe practices.
\end{abstract}


\end{frontmatter}

\section{Introduction}

Construction workers in the United States face fatality rates nearly four times the all-industry average \citep{bls2024}, and these rates persist despite advances in engineering controls and management systems \citep{newaz2024}. One explanation lies in how workers evaluate safety practices within organizational and situational constraints \citep{burns2013}. Knowing about a hazard or its control is not equivalent to acting on that knowledge. The difference often comes down to how workers weigh competing demands such as production pressure, physical comfort, peer expectations, and safety requirements \citep{swuste2012}. When confronted with such trade-offs, workers may judge the effort, time, or social cost of a safety practice as too high and bypass it as a result, a sequence that recurs as a driver of incidents \citep{fang2016}. These evaluations point to attitudes as critical determinants of whether safety practices are applied or bypassed \citep{hu2023}.

Behavioral theory supports this focus on attitudes. The Theory of Planned Behavior explains how beliefs inform attitudes, attitudes shape intentions, and intentions influence behavior \citep{ajzen1991}. Beliefs reflect factual acceptance, whereas attitudes are psychological tendencies expressed through situational evaluations of entities with favor or disfavor \citep{eagly1993}. In construction safety, workers form attitudes toward hazards, controls, rules, peers, and practices, and these evaluations determine whether workers value safety when trade-offs arise \citep{johari2020}. Assessing such attitudes therefore offers a practical lever for prevention. A clear understanding of how workers evaluate hazards, controls, and rules can sharpen the targeting of training, interventions, communications, and policy, and shift them from generic delivery toward the specific evaluations driving unsafe choices in a given workforce.

Measuring safety attitudes has attracted considerable attention in prior research, with primary methodologies relying on survey instruments that fall into two categories. The first asked workers to rate agreement with statements such as ``Sometimes it is necessary to take risks to get a job done'', assessing attitudes in generic settings \citep{kao2019}. The second asked workers to rate safety climate, inferring attitudes from how individual responses interact with organizational factors \citep{yuan2022}. These instruments typically draw on theoretical models, such as the Theory of Planned Behavior and Social Identity Theory, to establish construct validity. Depending on the model selected, researchers emphasize different facets of safety attitude, including attitudes toward risk \citep{kao2019}, hazard perception \citep{shin2014}, and protective effectiveness \citep{basahel2021}.

This survey-based inquiry has three notable limitations. First, although different theories emphasize distinct attitudinal dimensions and each contributes valid insight, the field has yet to integrate these dimensions into a coherent assessment structure. This fragmentation is widely recognized, with multiple studies calling for unified approaches \citep{kashmiri2020, loosemore2019, yuan2022, zhou2022}. Second, surveys typically elicit attitudes in general terms without accounting for the specific contexts in which those attitudes are situated, even though attitudes are known to shift with context \citep{cavazza2009}. Third, surveys are resource-intensive to distribute and analyze, which constrains how frequently they can be administered and leaves researchers and practitioners working from data that quickly grow stale. Without capturing both the full range of attitudinal dimensions, accounting for this situational variability, or producing timely measurements, surveys provide an incomplete picture of the attitudes that shape on-the-job decisions.

A promising yet underutilized alternative lies in naturalistic methods that capture attitudes through everyday communication \citep{dane2019}. Attitudes are observable in routine discourse, where workers frame safety-related practices as reasonable, excessive, necessary, or misguided \citep{shin2014}. Social media offers a particularly accessible data source, as workers voice attitudes through unsolicited posts, which helps mitigate the social desirability bias present in researcher-administered instruments \citep{kapoor2017}. The richness of such data is evident in \citet{yao2021}, who analyzed over 6,500 Twitter posts and found that construction workers frequently exchange safety opinions. Importantly, a single post can address multiple attitudinal dimensions simultaneously, and the timestamps attached to such posts enable temporal analysis of safety attitudes as they evolve. To date, however, no studies have investigated the potential of assessing multidimensional safety attitudes at the sector level using social media data.

A key barrier to assessing safety attitudes from naturalistic discourse is the absence of automated methods for systematically applying an attitudinal framework to large-scale text; manual analysis at this volume is prohibitive. Large Language Models (LLMs) now offer a methodological solution. LLMs can interpret contextual nuance and identify the stance a worker takes toward a safety practice \citep{thapa2025}. Recent studies demonstrate LLM effectiveness in extracting structured insights from unstructured construction safety text, including incident narratives and worker communications \citep{ahmadi2024, smetana2024}.

This study addresses two gaps. First, the field lacks a unified framework that integrates the attitudinal dimensions existing studies have examined in isolation. Second, it lacks an automated method for applying such a framework to naturalistic discourse at scale. To bridge these gaps, this study develops the Construction Safety Attitude Framework (CSAF), a multidimensional classification system, and operationalizes it through an LLM-based pipeline. Together, these contributions enable systematic attitude assessment across large volumes of worker communication. Three objectives guide the study:

\begin{itemize}
  \item Develop and validate a framework that captures the multidimensional nature of construction worker safety attitudes.
  \item Operationalize the framework through LLM-based classification, including prompt engineering and evaluation against expert coding.
  \item Apply the framework in a proof-of-value case study to identify attitudinal patterns across dimensions and safety topics in construction worker discourse.
\end{itemize}

Collectively, this work delivers a validated pipeline for capturing construction safety attitudes at scale using naturally occurring discourse. The framework and pipeline are designed for uptake by government agencies and industry associations to inform policy, targeted messaging, and training priorities, while remaining adaptable for employer-level use. By transforming unsolicited worker voice into a continuous, structured signal, this study establishes a scalable foundation for attitude-informed safety management across the sector.

\section{Literature Review}

The construction safety studies reviewed in this section were identified through a structured search of five databases: Scopus, Web of Science, EBSCO Academic Search Ultimate, Emerald Insight, and Google Scholar. Each was searched for peer-reviewed work published between 2000 and 2026, pairing construction-context terms (``construction worker,'' ``construction industry,'' ``construction site,'' ``construction safety'') with attitude-specific terms (``safety attitude,'' ``attitude toward safety,'' ``worker attitude,'' ``attitudinal''). Titles and abstracts were screened for direct relevance to construction worker safety attitudes, and the retained studies were grouped by the psychological theory each applied. Backward citation tracking on the reference lists of those studies identified the foundational works in which the theories originate.

Attitudes are psychological tendencies expressed through evaluating entities with favor or disfavor \citep{eagly1993}. Attitudes differ from beliefs, which represent factual acceptance, and from behaviors, which represent observable actions. This evaluative nature positions attitudes between what workers know or believe and what workers do \citep{ajzen1991}. Prior research has drawn on diverse psychological theories to examine safety attitude. A review of the theories applied and the dimensions they address highlights the need for integration.

Value Theory \citep{schwartz1992} provides a foundation for examining how workers evaluate and prioritize safety relative to competing demands such as productivity and cost. \citet{langford2000} observed that safety bonuses correlated with favorable attitudes toward safety management, while productivity bonuses correlated with pressure to prioritize production over safety. \citet{mccabe2005} reported that attitudes toward safety-productivity compatibility became more favorable with experience, suggesting that veteran workers come to view safety and efficiency as aligned rather than conflicting. \citet{wang2026} identified divergence across organizational levels, with managers holding more favorable prioritization attitudes than frontline workers. Together, these studies establish that principle integrity toward safety constitutes a distinct evaluative dimension of attitude shaped by incentives, experience, and organizational role.

The Theory of Planned Behavior \citep{ajzen1991} proposes that behavioral intention is shaped by attitude toward the behavior, subjective norms, and perceived behavioral control. Construction safety research has examined attitudes related to these factors. \citet{xu2018} examined attitudinal ambivalence, which arises when workers simultaneously hold positive and negative evaluations toward the same safety practice. Such ambivalence may stem from differences in context, underscoring the importance of examining attitudes within specific situations. \citet{pandit2018} demonstrated that crew-level cohesion and safety climate shape safety communication patterns measured through network density, indicating that workers' evaluations of safety form within the crew's communication structure rather than independently of it. Together, these studies establish attitudes toward social influence as a distinct evaluative dimension and show that such evaluations are situationally contingent rather than fixed.

Tripartite Attitude Theory \citep{rosenberg1960} proposes that attitudes comprise three components: cognitive, affective, and behavioral. \citet{loosemore2019} measured these components before and after safety training: cognitive attitudes improved, indicating that workers understood that safety mattered, while affective attitudes remained unchanged, meaning that workers still felt that safety measures were burdensome despite knowing their importance. This divergence demonstrates that attitude components can shift independently and that knowing safety matters does not guarantee feeling positively about safety practices. \citet{kashmiri2020} reported that workers with favorable attitudes identified more hazards than did workers with unfavorable attitudes. \citet{legishion2024} showed that attitudes predicted rule adherence. Collectively, these studies establish that cognitive and affective evaluations operate as distinct components with independent effects on hazard recognition and rule compliance.

Social Cognitive Theory \citep{bandura1986} provides constructs for examining attitudes toward personal capability to act safely and openness to learning from experience. \citet{hung2011} observed that workers held less favorable attitudes toward their own ability to control safety outcomes than did supervisors, and were more likely to attribute outcomes to luck rather than personal action. \citet{basahel2021} found that workers' attitudes toward their own ability to act safely predicted safety-related behavior in substation construction. \citet{tam2011} found that a mandatory training course shifted workers' safety attitudes, with completion producing greater attentiveness to safety, indicating that workers revise their evaluations in response to instructional feedback. Together, these studies establish self-efficacy and openness to learning as distinct evaluative dimensions, capturing how workers judge their own capability and how they adapt attitudes in light of their own and others' experience.

The Health Belief Model \citep{rosenstock1974} provides constructs for examining attitudes toward hazards and protective measures. \citet{shin2014} observed that workers held optimistic attitudes toward personal invulnerability at baseline, believing accidents were unlikely to happen to them. After witnessing accidents, attitudes shifted toward pessimism, but then gradually returned to optimism as memory faded and safer alternatives proved inconvenient. \citet{gharibi2016} reported that workers with accident experience held more favorable attitudes toward protective equipment than did workers without such experience, suggesting that direct consequences reshape how workers evaluate whether protections are worthwhile. Together, these studies establish attitudes toward hazard severity and protective effectiveness as distinct evaluative dimensions responsive to personal experience.

Social Identity Theory \citep{tajfel2004} provides constructs for examining attitudes toward collective responsibility. \citet{siu2003} established that favorable attitudes toward collective responsibility, defined as the beliefs that coworker safety is part of one's own duty, predicted safety outcomes at the site level. \citet{xu2023} modeled safety attitude resilience under disruption and observed that group-level factors drove the severity of attitude loss, while individual-level factors governed recovery. Collectively, these studies establish that attitudes toward collective responsibility constitute a distinct evaluative dimension linking individual orientations to group-level safety outcomes.

\begin{figure}[tb]
  \centering
  \includegraphics[width=\linewidth]{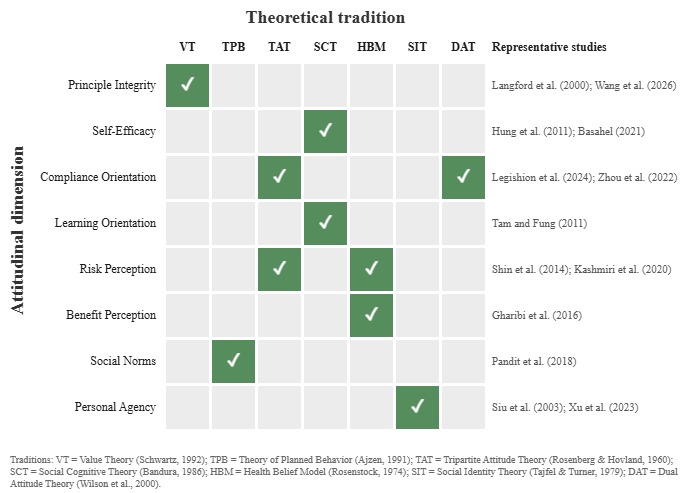}
  \caption{Coverage of attitudinal domains by theoretical tradition in prior construction safety research.}
  \label{fig:1}
\end{figure}

Beyond examining dimensions in isolation, some research has investigated how attitudinal dimensions interact. \citet{yuan2022} demonstrated that unfavorable attitudes combined with unfavorable climate perceptions produced unsafe behavior even when training and motivation were adequate, showing that attitudes operate in combination with other factors. Dual Attitude Theory \citep{wilson2000} contributes the distinction between deliberate and automatic evaluations that underlies how workers report their safety attitudes. \citet{zhou2022} applied Dual Attitude Theory and distinguished explicit attitudes, which are conscious evaluations workers can report, from implicit attitudes, which are automatic evaluations operating outside awareness. Their results indicated that implicit attitudes more strongly predicted behavior than explicit attitudes, and that attitudes affected compliance indirectly through motivation and knowledge rather than directly. Together, these studies suggest that attitudinal dimensions interact through multiple pathways. Fig.~\ref{fig:1} maps each theoretical tradition reviewed above to the attitudinal dimensions it has addressed, showing that coverage is distributed rather than unified. However, the absence of a systematic framework for organizing and integrating these dimensions has limited the field's ability to map their joint influence on safety behavior and to translate this understanding into targeted intervention.

\section{Construction Safety Attitude Framework}

\begin{figure}[tb]
  \centering
  \includegraphics[width=\linewidth]{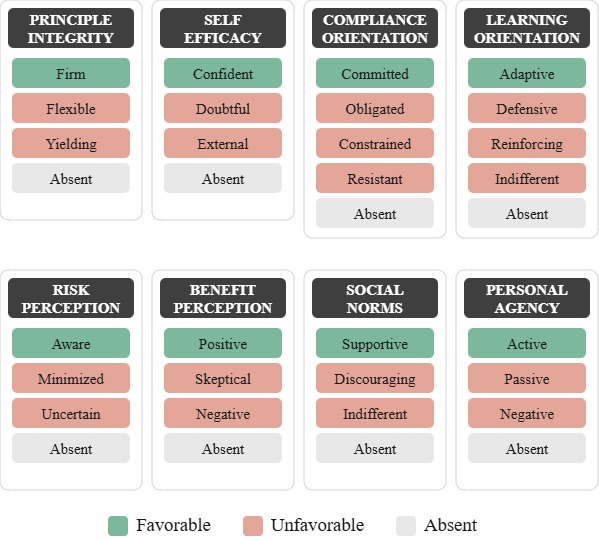}
  \caption{The Construction Safety Attitude Framework codebook: 34 codes across 8 dimensions, color-coded by valence (favorable, unfavorable, absent).}
  \label{fig:2}
\end{figure}

To address the theoretical fragmentation documented in Section~2, this study introduces the Construction Safety Attitude Framework (CSAF), which organizes attitudinal constructs into a unified classification system suited to naturalistic discourse. As shown in Fig.~\ref{fig:2}, the framework comprises eight dimensions synthesized from the constructs reviewed in the preceding section. Its theoretical structure (Section~3.1) and operational codes (Section~3.2) are presented below in their final form; however, both emerged from an iterative development and validation process detailed in Section~4 (Methods) and Section~5 (Results).

\subsection{Theoretical Structure}

CSAF organizes safety attitudes along eight dimensions, each defined by a distinct object toward which workers' evaluations are directed (Table~\ref{tab:1}). This object-centered structure is the organizing principle of the framework: it allows the dimensions to remain analytically distinct, with each capturing a discrete locus of evaluation relevant to on-site safety decisions.

\begin{table*}[tp]
\centering
\caption{Construction Safety Attitude Framework.}
\label{tab:1}
\small
\begin{tabular}{L{3.4cm} L{3.7cm} L{9.4cm}}
\toprule
\textbf{Dimension} & \textbf{Attitude Object} & \textbf{Definition} \\
\midrule
Principle Integrity & Safety priority & Evaluation of whether safety is maintained as a priority over production, income, or schedule \\
Self-Efficacy & Personal safety capability & Evaluation of personal ability to execute safety practices, use protective equipment, and control one's own safety outcomes \\
Compliance Orientation & Safety rules & Evaluation of whether safety rules deserve adherence based on intrinsic value or only under external enforcement \\
Learning Orientation & Prior experiences & Evaluation of openness to adapting behavior based on one's own or others' safety incidents and feedback \\
Risk Perception & Workplace hazards & Evaluation of whether specific hazards are personally relevant or exaggerated \\
Benefit Perception & Protective measures & Evaluation of whether safety controls provide genuine protection or are inadequate \\
Social Norms & Peer safety conduct & Evaluation of whether coworkers genuinely care about safety, discourage it, or remain indifferent \\
Personal Agency & Personal role in coworker safety & Evaluation of willingness to intervene for coworkers and engage in collective safety activities \\
\bottomrule
\end{tabular}
\end{table*}

Five dimensions are receptive, capturing how features of the work environment are perceived to act on the worker. Risk Perception and Benefit Perception concern the physical work environment, referring respectively to evaluations of workplace hazards and of the protective measures. Principle Integrity, Compliance Orientation, and Social Norms concern the social and organizational context, reflecting, respectively, evaluations of safety priority relative to production demands, the legitimacy of formal rules, and perceived peer conduct. Across these five dimensions, the evaluative question is how an external entity bears on the worker's safety and behavior.

Two dimensions are agentic, capturing how workers position their capability or role toward an external entity. Self-Efficacy evaluates perceived capability to control one's safety outcomes through safe work practices and protective equipment, projecting the self onto the physical work environment. Personal Agency evaluates perceived responsibility for and willingness to act on coworkers' safety, projecting the self onto the social environment. These two dimensions are central to understanding active safety conduct as distinct from passive response. The eighth dimension, Learning Orientation, is reflective. It evaluates the worker's own openness to adapting behavior in response to prior safety experiences, whether their own or others'. As the framework's only meta-attitudinal dimension, it captures the disposition through which other attitudes may be revised over time.

Together, these eight dimensions span the range of entities and orientations that shape on-site safety decisions, encompassing how workers are acted upon by their environment, how they project themselves toward it, and how they reflect on their own experience. Multiple evaluative targets may co-occur within a single communicative expression, a property that motivates the multi-label coding scheme developed in Section~3.2.

\subsection{Operational Codes}

To enable systematic coding of multidimensional safety attitudes in naturalistic discourse, each dimension is operationalized through discrete codes representing distinct evaluative orientations. The codebook comprises 34 codes: four codes each for six dimensions, and five each for Compliance Orientation and Learning Orientation. Codes are nominal categories without implied ordering.

The codebook initially used binary Favorable\,/\,Unfavorable categories, but this deductive scheme obscured meaningful distinctions within the same valence. For example, the statements ``I only wear my harness because they'll fire me'' and ``Harnesses are useless, I never wear one'' both express unfavorable orientations toward compliance, yet they reflect fundamentally different mechanisms: the first reflects external coercion (COMPLIANCE-OBLIGATED) while the second reflects active rejection (COMPLIANCE-RESISTANT). A binary scheme would collapse these two orientations into a single ``unfavorable'' category and obscure the distinct mechanisms driving each, limiting the framework's usefulness for diagnosing why a given attitude has formed. Through iterative refinement detailed in the Methods section, the binary categories were differentiated into the discrete codes presented in Table~\ref{tab:2}.

\begin{table*}[tp]
\centering
\caption{Construction Safety Attitude Framework Operational Codes.}
\label{tab:2}
\footnotesize
\begin{tabular}{L{2.7cm} L{4.2cm} L{3.7cm} L{5.4cm}}
\toprule
\textbf{Dimension} & \textbf{Guiding Question} & \textbf{Code} & \textbf{Definition} \\
\midrule
Principle Integrity & How does the worker prioritize safety against competing goals? & COMMITMENT-FIRM & Consistently prioritizes safety over competing goals \\
 & & COMMITMENT-FLEXIBLE & Prioritization varies by situation or context \\
 & & COMMITMENT-YIELDING & Prioritizes other goals over safety \\
 & & COMMITMENT-ABSENT & No value conflict discussed \\
\addlinespace
Self-Efficacy & How does the worker evaluate personal ability to control safety outcomes? & EFFICACY-CONFIDENT & Believes themselves capable of executing safety practices, using protective equipment, and staying safe \\
 & & EFFICACY-DOUBTFUL & Questions one's own ability to stay safe \\
 & & EFFICACY-EXTERNAL & Believes outcomes are beyond personal control \\
 & & EFFICACY-ABSENT & No capability assessment present \\
\addlinespace
Compliance Orientation & How does the worker evaluate safety rules? & COMPLIANCE-COMMITTED & Follows rules from internal conviction \\
 & & COMPLIANCE-OBLIGATED & Follows rules only under external pressure \\
 & & COMPLIANCE-CONSTRAINED & Wants to comply but faces barriers \\
 & & COMPLIANCE-RESISTANT & Actively chooses not to follow rules \\
 & & COMPLIANCE-ABSENT & No rule-following behavior discussed \\
\addlinespace
Learning Orientation & How does the worker respond to safety experiences? & LEARNING-ADAPTIVE & Open to changing behavior based on own or others' experiences \\
 & & LEARNING-DEFENSIVE & Defends current practice despite problems \\
 & & LEARNING-REINFORCING & Uses injury-free history to justify risky practices \\
 & & LEARNING-INDIFFERENT & Acknowledges experiences but shows no reflection or concern \\
 & & LEARNING-ABSENT & No response to safety experiences discussed \\
\addlinespace
Risk Perception & How does the worker evaluate workplace hazards? & RISK-AWARE & Acknowledges hazards as real and personally relevant \\
 & & RISK-MINIMIZED & Dismisses or downplays hazards \\
 & & RISK-UNCERTAIN & Shows genuine doubt about the level of danger \\
 & & RISK-ABSENT & No hazard evaluation present \\
\addlinespace
Benefit Perception & How does the worker evaluate protective measures? & BENEFIT-POSITIVE & Believes measures provide genuine protection \\
 & & BENEFIT-SKEPTICAL & Questions adequacy and seeks better options \\
 & & BENEFIT-NEGATIVE & Believes measures are ineffective or symbolic \\
 & & BENEFIT-ABSENT & No evaluation of safety measures \\
\addlinespace
Social Norms & How does the worker perceive coworkers' safety behavior? & NORMS-SUPPORTIVE & Perceives peers as genuinely supporting safe practices \\
 & & NORMS-DISCOURAGING & Perceives peers as discouraging or mocking safety \\
 & & NORMS-INDIFFERENT & Perceives peers as indifferent or unconcerned about safety \\
 & & NORMS-ABSENT & No evaluation of peer behavior mentioned \\
\addlinespace
Personal Agency & How does the worker view their personal role in coworkers' safety? & AGENCY-ACTIVE & Willing to intervene and help coworkers stay safe \\
 & & AGENCY-PASSIVE & Views coworker safety as outside personal role \\
 & & AGENCY-NEGATIVE & Actively undermines safe practices of others \\
 & & AGENCY-ABSENT & No evaluation of role in others' safety discussed \\
\bottomrule
\end{tabular}
\end{table*}

\section{Methods}

\begin{figure}[tb]
  \centering
  \includegraphics[width=\linewidth]{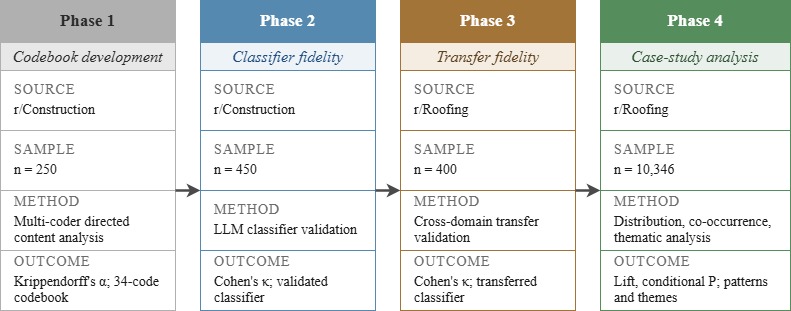}
  \caption{Four-phase methodological workflow showing the data source, sample, method, and outcome at each stage.}
  \label{fig:3}
\end{figure}

This section describes the methods by which the CSAF was operationalized, validated, and applied through a four-phase process. Phase 1 developed the CSAF codebook through iterative human coding of construction discourse. Phase 2 operationalized the codebook as an automated LLM classifier and evaluated its performance against human consensus on an independent sample of social media data. Phase 3 tested whether classifier performance held when the classifier transferred to a different construction subdomain. Phase 4 applied the classifier to the full transfer corpus and analyzed the resulting patterns through quantitative and thematic analyses, both to test construct validity and to demonstrate practical value of the CSAF. Fig.~\ref{fig:3} presents the four-phase workflow, specifying data source, sample, method, and outcome at each stage. The study was approved by the Texas A\&M University Institutional Review Board.

\subsection{Data Source}

Reddit was selected as the discourse source for three reasons. First, its pseudonymous posting structure allows workers to discuss safety without fear of employer retaliation or social judgment, mitigating social desirability bias. Second, its threaded conversation structure preserves the sequential context of discourse \citep{kapoor2017}, allowing each comment to be interpreted alongside the preceding conversation. Third, Reddit is organized into subreddits, user-created topical communities whose members exchange views on a shared subject. In this study, data were collected from two subreddits (r/Construction and r/Roofing) to support both initial development and transfer validation of the framework. r/Construction, a broad community spanning all construction trades, provided data for codebook development (Phase 1) and LLM classifier validation (Phase 2). r/Roofing, a specialized trade community, provided data for transfer validation (Phase 3) and case-study application (Phase 4). r/Roofing was selected for the case study because it ranks among the most hazardous construction specialties \citep{bls2024}, and because its trade-specific terminology and regulatory frameworks provide a rigorous test of whether a codebook developed on general construction discourse transfers to a specialized subdomain.

\subsubsection{Data Collection}

\begin{figure}[tb]
  \centering
  \includegraphics[width=\linewidth]{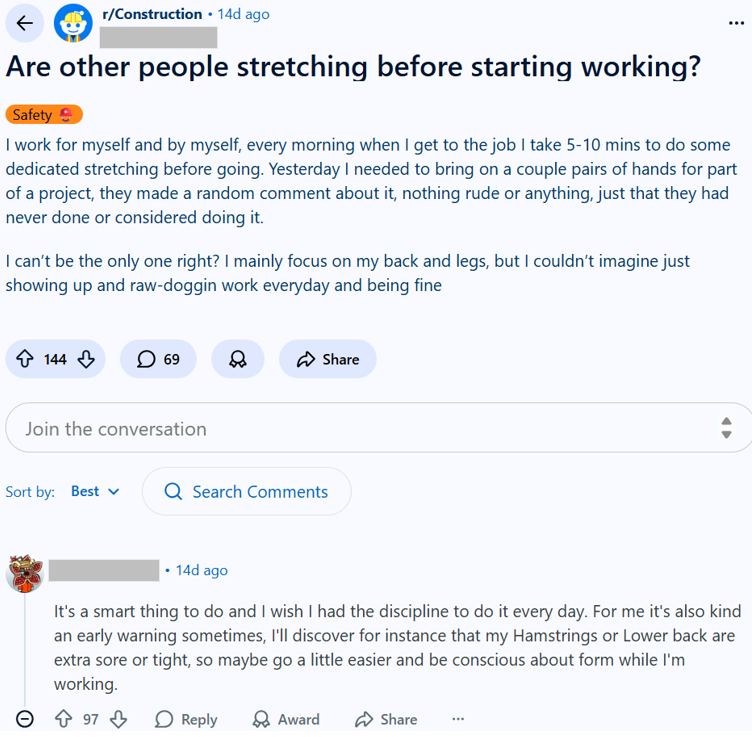}
  \caption{Example safety-flaired post from r/Construction with a threaded comment. The safety flair (green tag, top left) is the moderator-enforced topic label used to collect the r/Construction dataset.}
  \label{fig:4}
\end{figure}

All data were retrieved through the Reddit API. The two subreddits required different collection strategies because r/Construction maintains moderator-enforced topic labels and r/Roofing does not. For Phases 1 and 2, r/Construction posts and comments from December 2025 through February 2026 were collected using the subreddit's safety flair, a topic label that posters attach to classify their submissions as safety-oriented. For Phases 3 and 4, r/Roofing posts and comments from September 2016 through February 2026 were collected. Because r/Roofing has no equivalent flair, a Boolean keyword search isolated posts containing safety-related terms (for example, ``fall,'' ``harness,'' ``OSHA,'' ``PPE,'' ``injury,'' ``hazard,'' ``anchor,'' or ``tie-off''). Each data point was a single post or comment, and each comment was analyzed alongside its preceding thread so that evaluative expressions could be interpreted in their conversational context. Fig.~\ref{fig:4} illustrates this structure with an example safety-flaired r/Construction post and a threaded comment: the post expresses an attitude toward a protective practice, and the comment reveals an evaluative stance formed in response. No personally identifiable information was collected.

\subsubsection{Pre-Processing and Authenticity Screening}

\begin{figure}[tb]
  \centering
  \includegraphics[width=\linewidth]{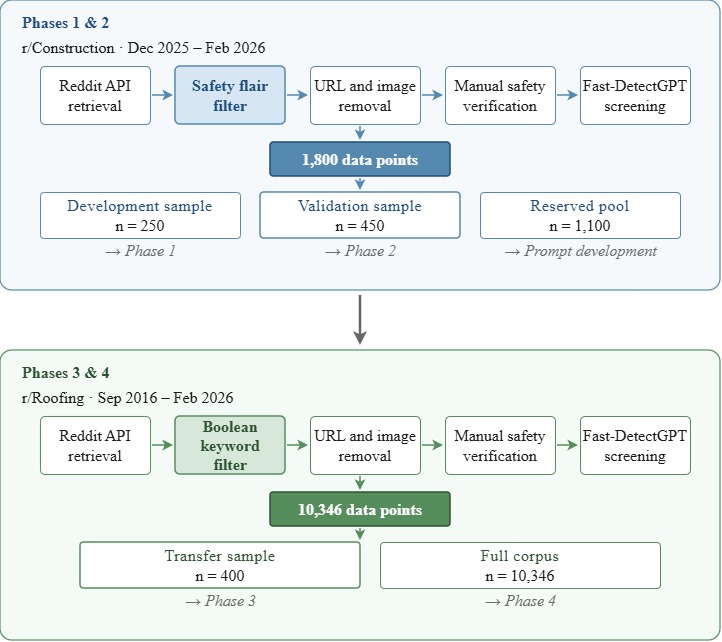}
  \caption{Data collection and filtering pipeline for the codebook-development corpus (r/Construction) and the case-study corpus (r/Roofing).}
  \label{fig:5}
\end{figure}

Each dataset underwent the same pre-processing and authenticity screening pipeline. Uniform resource locators (URLs) and image links were removed. All collected data points were then manually reviewed by trained coders at the individual post level to verify safety relevance, operationally defined as topical alignment with an industry association safety manual. Ultimately, 97.4\% of r/Construction (Phases 1 and 2) and 96.5\% of r/Roofing (Phases 3 and 4) data points were confirmed as containing safety-related discourse, with the remainder excluded. Reddit's platform-level safeguards, including AutoModerator rule-based filtering and mandatory bot-account labeling \citep{jhaver2019}, reduce spam and automated posting. To detect AI-generated text that mimics human authorship \citep{lloyd2025}, both datasets were then screened using Fast-DetectGPT \citep{bao2023}, a zero-shot detector that scores each text on how closely its token-probability patterns resemble language-model output. Texts scoring above 0.5 were classified as AI-generated, a threshold that flags clear cases while limiting false positives on casual discourse \citep{jung2025}. A 50-token minimum was enforced for reliable scoring \citep{wei2025}. No AI-generated content was identified in either dataset. Fig.~\ref{fig:5} presents the data collection and filtering pipeline.

\subsubsection{Final Datasets and Sampling}

The pipeline yielded 1,800 data points from r/Construction (120 posts and 1,680 comments) and 10,346 from r/Roofing (296 posts and 10,050 comments). Following established content-analysis sampling practice for digital corpora \citep{daniel2023, lacy2015}, the r/Construction set was partitioned by simple random sampling into three non-overlapping samples. A development sample ($n = 250$) was constructed for Phase 1 codebook elaboration. This sample size meets thematic-saturation thresholds reported in the qualitative methodological literature \citep{guest2006, hennink2017} and is sufficient to accommodate the lexical and contextual heterogeneity of subreddit discourse and to support stable code refinement \citep{hsieh2005, krippendorff2018, miles2015}. A validation sample ($n = 450$) was sized using Cochran's formula for proportion estimation ($Z = 1.96$, $p = 0.5$, $e = 0.05$), which yields a minimum of 384 observations for 95\% confidence with a $\pm 5\%$ margin of error around classifier performance metrics \citep{qing2025}; this approach follows recommended practice for content-analysis reliability sampling \citep{lacy2015, lacy1996}, with the additional cases providing precision for class-level estimates and for kappa-based inter-rater agreement \citep{sim2005}. A reserved pool ($n = 1{,}100$) was retained for iterative prompt development and exemplar selection. The non-overlap ensured that no data point used to develop the codebook was later used to evaluate the classifier. The full r/Roofing set supported transfer validation in Phase 3 and case-study application in Phase 4, consistent with hybrid manual-computational content-analysis designs for large digital corpora \citep{lewis2013}.

\subsection{Phase 1: Codebook Development}

Phase 1 translated the eight CSAF dimensions derived from the theoretical foundations in Section~2 into the 34 discrete codes listed in Table~\ref{tab:2}, using the r/Construction development sample ($n = 250$). In parallel, each data point was also classified by safety topic using an adapted version of the National Roofing Contractors Association (NRCA) safety manual chapters taxonomy. This dual classification supported subsequent analysis of how attitudes vary across safety topics.

The codebook followed directed content analysis \citep{hsieh2005}, an approach suited to CSAF because the dimensions were theoretically grounded, while the specific codes within each dimension needed to emerge from the data. Two doctoral researchers with training in qualitative content analysis and construction safety research independently coded all data points, each assigning one code per attitudinal dimension and one or more safety topic classifications while viewing the data point alongside its preceding thread for context. Disagreements were resolved through discussion; unresolved cases were adjudicated by a third coder, a Certified Safety Professional with expertise in safety attitude measurement. This same three-coder configuration was used in all subsequent human-coding phases.

Coding proceeded in iterative rounds. Coders first classified 50 randomly selected data points using the three-category scheme (Favorable, Unfavorable, Absent) \citep{campbell2013}. Initial reliability across this batch ranged from $\alpha = 0.45$ to $0.62$, indicating that the three-category scheme did not yet achieve acceptable agreement. Analysis of disagreement patterns revealed that the three initial categories obscured meaningful distinctions within the same valence, prompting expansion from three categories to the 34-code scheme. Per-dimension $\alpha$ values were aggregated into a single overall estimate using engagement weighting. Each dimension was weighted by the count of data points where it was engaged, defined as at least one coder assigning a non-ABSENT code. This weighting ensured that dimensions workers more often addressed contributed proportionally more to the aggregate. Consistency was then checked on a separate subset of 50 data points drawn from the remaining unanalyzed portion of the development sample. Coding continued through additional batches until theoretical saturation was achieved, operationally defined as three consecutive batches of 50 data points yielding no new codes \citep{guest2006}.

During iterative coding, recurring edge cases required explicit resolution to maintain consistency across coders and rounds. Five decision rules were established through coder discussion during Phase 1 and applied uniformly across all subsequent phases. First, sarcasm was identified through contextual cues such as inconsistency between literal content and surrounding framing, and coded by the sincere attitudinal position implied by context. Second, rhetorical questions were treated as declarative statements of the implied position. Third, attitude evolution within a single data point was coded by the final expressed state. Fourth, reported speech was coded by the author's stance toward the reported attitude, not by the attitude of the reported speaker. Fifth, ABSENT was applied when a dimension was not addressed and was treated as the absence of evaluative expression rather than a neutral attitude.

\subsection{Phase 2: LLM Classifier Fidelity}

The 34-code scheme finalized in Phase 1 produced reliable human classifications, but manual coding cannot scale to the thousands of data points required for the case study. Phase 2 tested whether an LLM could replicate human coding at the accuracy needed for large-scale application \citep{grimmer2022}.

\subsubsection{Ground Truth Establishment}

The validation sample comprised the 450 data points reserved for establishing ground truth from the r/Construction dataset. This sample was separate from the 250-point development sample used in the codebook-development phase (Phase 1). The same three-coder configuration from Phase 1 independently classified all 450 data points using the finalized codebook. Disagreements were resolved through discussion, and unresolved cases were adjudicated by the third coder. The resulting consensus codes were the ground truth against which LLM output was compared.

\subsubsection{Prompt Engineering and LLM Selection}

The LLM classifier was developed through iterative prompt engineering using the TIDD-EC framework (Task, Identity, Data, Definitions, Edge cases, Constraints; \citealp{vivas2024}). Refinement was conducted on 50 held-out data points drawn from the reserved pool ($n = 1{,}100$) that were used exclusively for prompt development. The prompt mirrored the human coding process from the codebook-development phase (Phase 1), with each of the six TIDD-EC components carrying forward a specific element of that process into the automated pipeline.

\begin{figure}[tb]
  \centering
  \includegraphics[width=\linewidth]{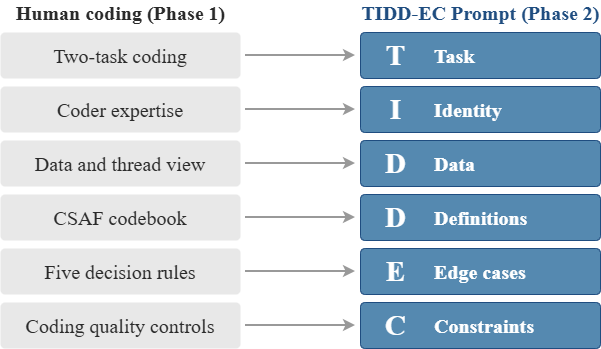}
  \caption{Task, Identity, Data, Definitions, Edge cases, and Constraints (TIDD-EC) prompt framework.}
  \label{fig:6}
\end{figure}

\begin{figure}[tb]
  \centering
  \includegraphics[width=\linewidth]{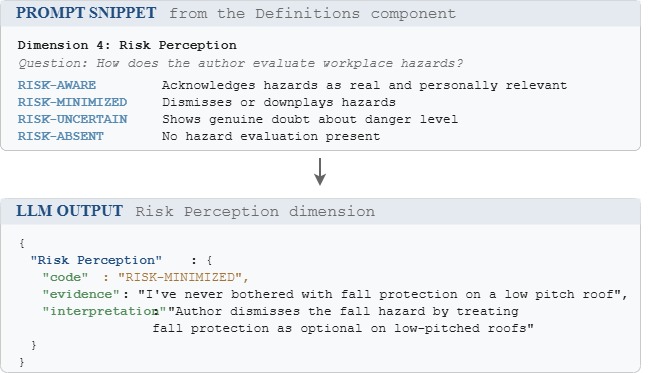}
  \caption{Prompt and output example for the Risk Perception dimension. Tag definitions (top) and the resulting classification of one data point (bottom).}
  \label{fig:7}
\end{figure}

Task definition established two classification objectives: assign one code per CSAF dimension for each data point across all eight attitudinal dimensions, and classify the data point by safety topic, with multiple topic assignments permitted when a data point addressed more than one safety domain. Identity prompting positioned the model as an expert safety-psychology analyst trained in qualitative content analysis. Data input specified the format for receiving Reddit posts and comments with their preceding thread context. Definitions embedded the full CSAF codebook from Table~\ref{tab:2} with the guiding questions and definitions from Section~3. Edge cases encoded the five decision rules from Phase 1. Constraints specified required behaviors (extract exact quotes as evidence, analyze all dimensions, tag the author's final expressed attitude) and prohibited behaviors (paraphrasing evidence, skipping dimensions, assigning codes without evidence; \citealp{sammour2026}). Constraints also specified the output format: structured JSON with assigned code, verbatim evidence, and an interpretation linking evidence to code. Fig.~\ref{fig:6} presents the TIDD-EC prompt framework, with each component mapped to its source in the human coding process. Fig.~\ref{fig:7} illustrates how the framework is operationalized in the actual prompt and structured output, using the Risk Perception dimension as a worked example.

GPT-OSS-120B, an open-weight reasoning model with 117 billion parameters, was selected for all LLM classification. An open-weight architecture was preferred because it allows full reproducibility. Temperature was set to 0.7. Classification stability at this non-deterministic setting was tested by processing 50 data points three times and quantifying the proportion of (data point $\times$ dimension) cells that produced identical codes across the three runs, with $\geq 0.90$ agreement adopted as the threshold for acceptable reproducibility.

\subsubsection{Agreement Metrics}

Agreement between LLM output and human consensus was quantified per dimension using Cohen's kappa \citep{cohen1960}, with $\kappa \geq 0.70$ adopted as the minimum threshold \citep{krippendorff2018, landis1977}. Raw classification accuracy was not used as the primary metric because the high prevalence of ABSENT codes across dimensions would inflate accuracy percentages; Cohen's kappa and F1 scores provide chance-corrected and class-balanced performance measures. To characterize error patterns beyond aggregate agreement, precision, recall, and F1 were also computed at the code level within each dimension, and confusion matrices were inspected at the dimension level. Per-dimension $\kappa$ values were aggregated using the same engagement-weighting introduced for $\alpha$ in Phase 1, with engagement defined here as Consensus or LLM assigning a non-ABSENT code. Per-code precision, recall, and F1 within each dimension were weighted by support, defined as the count of consensus instances of the code, and then averaged across dimensions. Both weighting choices follow standard practice for multi-class classification with unbalanced class frequencies \citep{grandini2020, sokolova2009} and report agreement on the dimensions and codes that carry analytical weight rather than treating each dimension or code equally regardless of prevalence.

\subsection{Phase 3: Transfer Fidelity Testing}

Phase 3 tested whether the classifier fidelity established in Phase 2 held on roofing-specific language and trade jargon. A classifier developed on a single data source risks overfitting to that source's vocabulary and discourse patterns. This concern, known as domain shift, is well documented in natural language processing \citep{ramponi2020}. Transfer testing on an independent trade community therefore assesses whether the classifier generalizes across construction subdomains. Roofing discourse presents a strong test because it uses trade-specific terminology (for example, ``flashing,'' ``deck,'' ``square'') that carries safety-relevant meaning different from general construction usage.

A random sample of 400 data points was drawn from the r/Roofing dataset for transfer validation. This sample size provided 95\% confidence with $\pm 5\%$ margin of error \citep{sim2005}. The classification prompt from Phase 2 was applied without modification, testing the classifier under identical conditions to Phase 2 except for the change in domain. The three-coder protocol from Phases 1 and 2 was applied. Agreement between the LLM and the reconciled human consensus was quantified using Cohen's kappa per dimension with the same $\kappa \geq 0.70$ threshold as Phase 2, and intercoder reliability was quantified using Krippendorff's alpha with the same $\alpha \geq 0.67$ threshold as Phase 1.

\subsection{Phase 4: Case Study Application}

Phase 4 applied the classifier to the full r/Roofing dataset ($n = 10{,}346$). The case study served two purposes. First, it extended the framework testing from measurement quality, established in the prior phases, to construct validity, defined as the extent to which a measurement instrument captures the theoretical construct it is designed to measure \citep{messick1995}. Second, it demonstrated practical value by showing that CSAF enables meaningful patterns to emerge from worker discourse. The analysis took two forms: the distribution of attitudes across dimensions and safety topics, and the co-occurrence of codes across dimensions.

\subsubsection{Attitude Distributions}

Attitude distributions were characterized using three complementary quantities: engagement rate, code share, and the favorable-to-unfavorable ($F/U$) ratio. Each quantity was computed for the full dataset and separately for each safety topic to test whether safety domains differed in their attitudinal patterns. The engagement rate measured how often each dimension was addressed in the discourse. For dimension $d$, the engagement rate ($E_d$) was computed as:

\begin{equation}
E_d = \frac{n_d}{N}
\end{equation}

where $n_d$ is the number of data points with a non-ABSENT code on dimension $d$, and $N = 10{,}346$ is the total number of data points. A high engagement rate indicates that workers frequently address the dimension; a low engagement rate indicates that the dimension rarely surfaces in worker discourse.

The code share ($S_{c,d}$) measured the prevalence of each specific code within the engaged portion of a dimension for a given safety topic. For code $c$ on dimension $d$, the code share was computed as:

\begin{equation}
S_{c,d} = \frac{n_{c,d}}{n_d}
\end{equation}

where $S_{c,d}$ is the code share of code $c$ on dimension $d$, and $n_{c,d}$ is the number of data points assigned code $c$ on dimension $d$ ($n_d$ is defined in Equation~1). The $S_{c,d}$ values sum to 1.0 across the codes of a dimension and are equivalent to the conditional probability $P(c \mid \text{non-ABSENT on } d)$. Engagement rate and code share capture distinct properties of the data: how often a dimension was addressed, and how attitudes were distributed when it was.

The favorable-to-unfavorable ($F/U$) ratio summarized the attitudinal balance of each dimension for a given safety topic. For dimension $d$, the ratio was computed as:

\begin{equation}
FU_d = \frac{f_d}{u_d}
\end{equation}

where $FU_d$ is the $F/U$ ratio for dimension $d$, $f_d$ is the count of favorable codes on dimension $d$, and $u_d$ is the count of unfavorable codes on dimension $d$. Values above 1.0 indicate favorable dominance on the dimension, values below 1.0 indicate unfavorable dominance, and values near 1.0 indicate balance. Codes assigned to each valence are listed in Fig.~\ref{fig:2}.

\subsubsection{Attitude Co-Occurrence}

Intercoder reliability (Phase 1) and classifier fidelity (Phases 2 and 3) tested whether CSAF can be applied consistently across human coders and the automated classifier. The analysis of attitude co-occurrence addressed a separate measurement property: construct validity. For a multidimensional framework, construct validity requires that the dimensions measure related yet distinct attitudinal constructs. Two measures from association rule analysis, pairwise lift \citep{brin1997} and conditional probability \citep{hahsler2005}, were used to test this property across the 26 active codes in the codebook.

First, pairwise lift compared how often two codes co-occurred in the data to how often they would co-occur by chance alone. For any two codes $A$ and $B$, lift was computed in Equation~4 as:

\begin{equation}
\mathrm{Lift}(A, B) = \frac{P(A \cap B)}{P(A) \times P(B)}
\end{equation}

where $P(A \cap B)$ is the probability that codes $A$ and $B$ both appear in the same data point, and $P(A)$ and $P(B)$ are the probabilities that codes $A$ and $B$ each appear in any data point. The denominator represents the chance baseline. A lift value above 1.0 indicates codes co-occur more than chance, a value below 1.0 indicates they co-occur less than chance, and a value near 1.0 indicates independence. Following conventions in association rule analysis \citep{hahsler2017, tan2004}, lift values at or above 2.0 were treated as evidence of co-occurrence above chance, the band 0.80 to 1.25 was treated as approximate independence, and values below 0.80 were treated as suppression. Pairs that never co-occurred yielded undefined lift values and were reported separately as instances of mutual exclusion.

Second, conditional probability resolved the directional ambiguity in pairwise lift by quantifying how often one code appeared given that another was present. For codes $A$ and $B$, conditional probability was computed in Equation~5 as:

\begin{equation}
P(B \mid A) = \frac{P(A \cap B)}{P(A)}
\end{equation}

where $P(B \mid A)$ is the probability that code $B$ appears given that code $A$ is present, $P(A \cap B)$ is the probability that codes $A$ and $B$ both appear in the same data point, and $P(A)$ is the probability that code $A$ appears in any data point. Conditional probability is directional. $P(B \mid A)$ and $P(A \mid B)$ need not be equal, and the asymmetry between them reveals the structure of the association. Codes that travel together in both directions function as near-duplicates and threaten construct validity, while codes for which one direction is high and the other substantially lower indicate hierarchical structure. Following psychometric conventions for high inter-item correlation in scale development \citep{hair2019}, conditional probabilities exceeding 0.80 in both directions for the same code pair were treated as evidence of redundancy.

\subsubsection{Qualitative Illustration of Unfavorable Attitudes}

Thematic analysis of code-filtered subsets complemented the quantitative analyses by characterizing the substantive content of unfavorable attitudes within specific safety topics. The topic and dimensions for thematic analysis were selected on engagement: the most-engaged safety topic and the CSAF dimensions most engaged within that topic. Each subset comprised the data points carrying unfavorable codes on the selected CSAF dimensions within the selected safety topic. The three-coder team analyzed each data point alongside its preceding thread context and applied thematic analysis \citep{clarke2017} to identify recurring themes in the reasoning, framing, and circumstances workers expressed. Themes were generated inductively through cross-data-point analysis and reported with their definitions, frequencies, and verbatim exemplar quotes alongside the quantitative results.

\section{Results}

\subsection{Reliability}

Table~\ref{tab:3} presents reliability results for both the CSAF codebook and the LLM classifier. Codebook quality was established through strong inter-rater agreement among all samples. On the 250-point r/Construction development sample used in Phase 1, engagement-weighted Krippendorff's $\alpha = 0.85$ across the eight CSAF dimensions, above the 0.67 threshold on every dimension. Human coder agreement held on the 450-point r/Construction validation sample used in Phase 2 ($\alpha = 0.86$) and on the 400-point r/Roofing transfer sample used in Phase 3 ($\alpha = 0.81$). The LLM classifier demonstrated strong performance in operationalizing the codebook. GPT-OSS-120B reproduced human consensus on the validation sample at engagement-weighted Cohen's $\kappa = 0.90$, above the 0.70 threshold on every dimension, with weighted precision 0.98, recall 0.98, and F1 0.98 across the eight dimensions. Performance held on the r/Roofing transfer sample at $\kappa = 0.89$, with weighted precision 0.98, recall 0.97, and F1 0.97 across the eight dimensions. Classifier output was stable across three independent runs on the same 50 data points (388 of 400 cells identical, 0.97 agreement).

\begin{table*}[tp]
\centering
\caption{Measurement quality and data quality findings.}
\label{tab:3}
\small
\begin{tabular}{L{2.7cm} L{3.9cm} L{3.4cm} C{2.0cm} L{4.2cm}}
\toprule
\textbf{Category} & \textbf{Property} & \textbf{Measure} & \textbf{Threshold} & \textbf{Result} \\
\midrule
Coder reliability & Codebook-development sample ($n = 250$) & Krippendorff's $\alpha$ & $\geq 0.67$ & 0.85 \\
 & Validation sample ($n = 450$) & Krippendorff's $\alpha$ & $\geq 0.67$ & 0.86 \\
 & Transfer sample ($n = 400$) & Krippendorff's $\alpha$ & $\geq 0.67$ & 0.81 \\
\addlinespace
Classifier fidelity & r/Construction validation ($n = 450$) & Cohen's $\kappa$ & $\geq 0.70$ & 0.90 \\
 & r/Construction validation ($n = 450$) & Precision, recall, F1 & --- & $P = 0.98$; $R = 0.98$; F1 $= 0.98$ \\
 & r/Roofing transfer ($n = 400$) & Cohen's $\kappa$ & $\geq 0.70$ & 0.89 \\
 & r/Roofing transfer ($n = 400$) & Precision, recall, F1 & --- & $P = 0.98$; $R = 0.97$; F1 $= 0.97$ \\
\addlinespace
Classification stability & Repeated runs (400 cells) & Identical codes across 3 runs & $\geq 0.90$ & 0.97 \\
\bottomrule
\end{tabular}
\end{table*}

\subsection{Content and Construct Validity}

The CSAF codebook demonstrated content validity across two criteria (Table~\ref{tab:4}). The codebook reached theoretical saturation when three consecutive coding batches yielded no new codes during codebook development, confirming that the codebook captured the full range of attitudinal categories present in the development corpus. All 26 active codes appeared at least once when the codebook was applied to the case-study corpus, confirming that no code represents an artificial category.

\begin{table*}[tp]
\centering
\caption{Content-validity and construct-validity findings.}
\label{tab:4}
\small
\begin{tabular}{L{2.5cm} L{2.9cm} L{3.6cm} C{2.0cm} L{5.4cm}}
\toprule
\textbf{Category} & \textbf{Property} & \textbf{Measure} & \textbf{Threshold} & \textbf{Result} \\
\midrule
Content validity & Codebook saturation & Consecutive batches with no new codes & $\geq 3$ batches & 3 consecutive; no new codes \\
 & Empirical code coverage & CSAF active codes present in case-study data & 26 of 26 & 26 of 26 active codes present \\
\addlinespace
Construct validity & Relatedness & Lift distribution across 254 defined pairs & Concentrated above 1.0 & $165 \geq 2.0$; 39 in 1.25--2.0; 27 in 0.80--1.25; $23 < 0.80$; mean 4.57 \\
 & Distinctness & Code pairs with $P(B \mid A)$ and $P(A \mid B)$ both $> 0.80$ & 0 pairs & 0 of 254 defined pairs \\
\bottomrule
\end{tabular}
\end{table*}

\begin{figure}[tb]
  \centering
  \includegraphics[width=\linewidth]{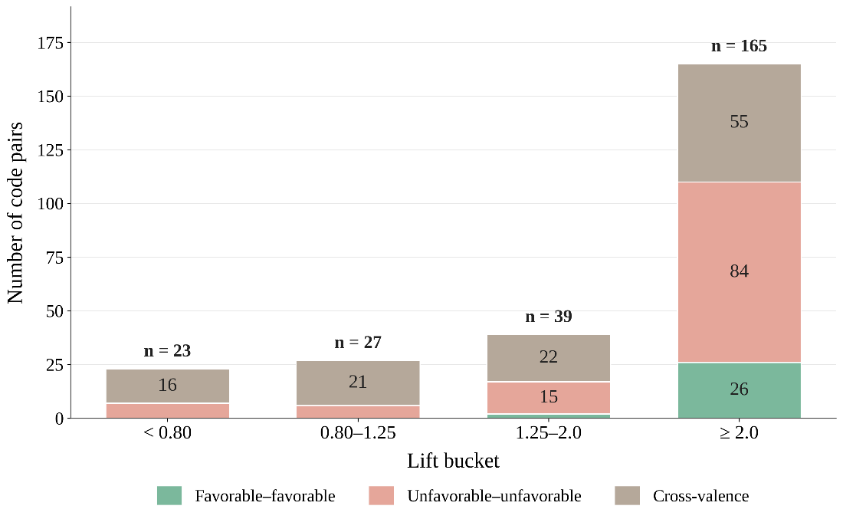}
  \caption{Lift distribution across the 254 defined code pairs, stacked by valence pattern.}
  \label{fig:8}
\end{figure}

\begin{figure*}[tp]
  \centering
  \includegraphics[width=\linewidth]{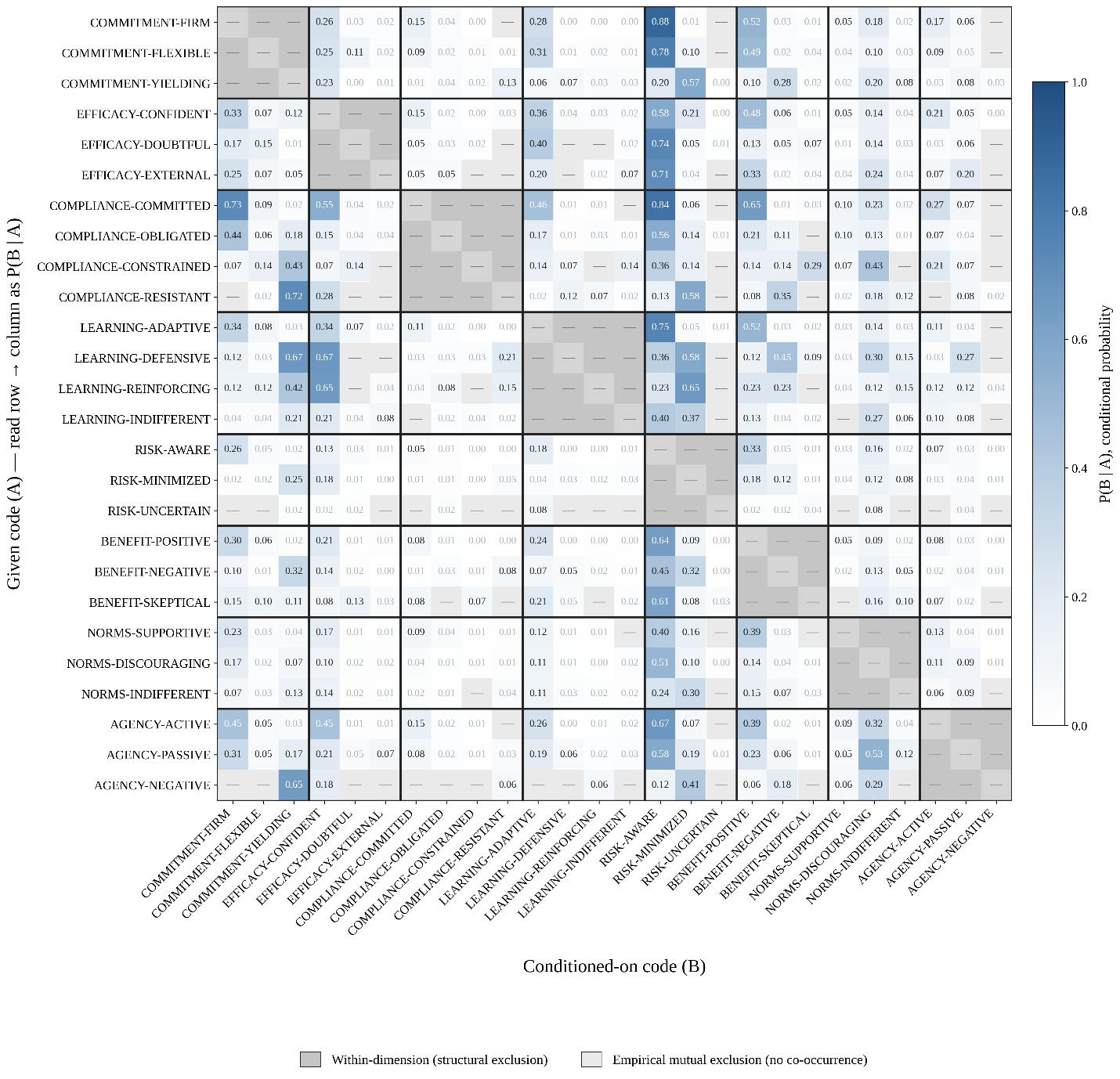}
  \caption{Conditional probability matrix across codes. Rows are conditioning codes ($A$); columns are the conditioned-on codes ($B$); cell values show $P(B \mid A)$.}
  \label{fig:9}
\end{figure*}

The CSAF codes demonstrated construct validity as related but distinct dimensions of safety attitudes, assessed through pairwise lift and conditional probability (Table~\ref{tab:4}). Lift was defined for 254 pairs; the 41 pairs that never co-occurred were excluded from the analysis. Lift distribution established that the codes are related. As shown in Fig.~\ref{fig:8}, the distribution concentrated in the high-lift bands rather than spreading across the range: of the 254 defined pairs, 165 (65.0\%) showed lift at or above 2.0, indicating that these code combinations appeared together at least twice as often as independence would predict. Mean lift was 4.57 and median 2.85, both well above the value of 1.0 that would indicate statistical independence. Conditional probability established that the codes are distinct. As shown in Fig.~\ref{fig:9}, no defined pair exhibited both $P(B \mid A) > 0.80$ and $P(A \mid B) > 0.80$, the threshold for redundancy adopted from psychometric scale-development conventions. This confirms that none of the 26 active codes is statistically interchangeable with another and that each code does measurement work the others do not.

The lift distribution also revealed meaningful co-occurrence patterns (Fig.~\ref{fig:8}). Pairs of unfavorable codes co-occurred at a mean lift of 6.90, the highest among all pairing types, suggesting that workers expressing one unfavorable safety attitude tend to express multiple unfavorable attitudes simultaneously in natural discourse. Furthermore, pairs combining one favorable and one unfavorable code produced a mean lift of 2.32 across 114 such pairs, reflecting the attitudinal complexity. Workers frequently expressed favorable attitudes toward one safety dimension while simultaneously expressing unfavorable attitudes toward another, a pattern that unidimensional measures would obscure and that underscores the necessity of treating safety attitudes as inherently multidimensional.

\subsection{Case Study Findings}

The validated LLM classifier was applied to the full r/Roofing corpus ($n = 10{,}346$) to demonstrate the practically actionable patterns that CSAF produces.

\subsubsection{Safety Attitude Distributions}

\begin{figure}[tb]
  \centering
  \includegraphics[width=\linewidth]{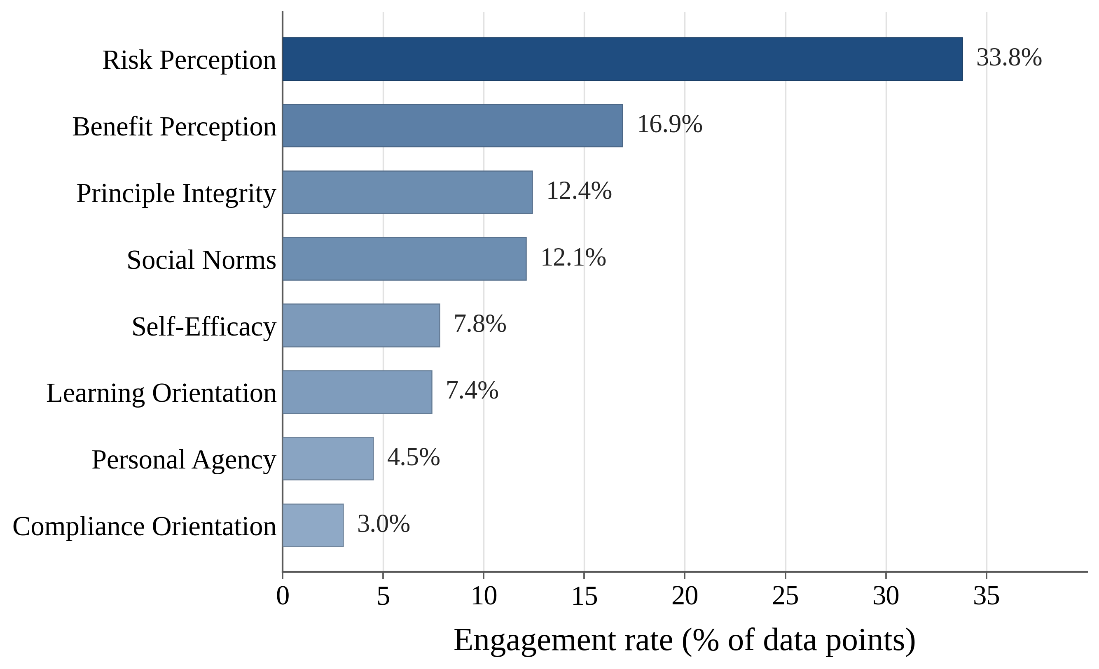}
  \caption{Engagement rate per CSAF dimension.}
  \label{fig:10}
\end{figure}

Engagement captured how frequently each attitudinal dimension surfaced in discourse. Attitudes were unevenly distributed across the eight CSAF dimensions, ranging from Risk Perception as the most engaged (33.8\%) to Compliance Orientation as the least (3.0\%) (Fig.~\ref{fig:10}). Engagement concentrated on the physical work environment, with Risk Perception and Benefit Perception together accounting for over half of all engaged comments. Dimensions reflecting organizational and social factors, including Principle Integrity and Social Norms, followed at slightly lower engagement levels.

\begin{figure}[tb]
  \centering
  \includegraphics[width=\linewidth]{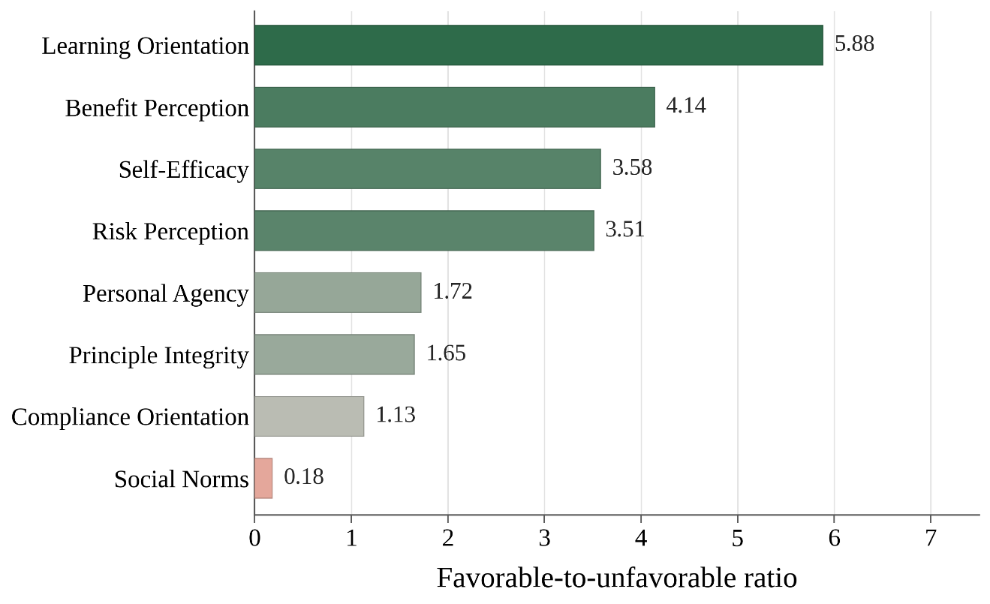}
  \caption{Favorable-to-unfavorable ($F/U$) ratio per CSAF dimension.}
  \label{fig:11}
\end{figure}

The $F/U$ ratio, computed as the proportion of favorable to unfavorable expressions, identified which valence prevailed when the dimension was engaged (Fig.~\ref{fig:11}). Six dimensions skewed favorable, ranging from Learning Orientation as the most favorable ($F/U = 5.88$) to Social Norms as the least ($F/U = 0.18$). Social Norms was the only dimension that skewed unfavorable overall, indicating that workers in natural discourse characterized their peer environment as discouraging of safety behavior far more often than as supportive of it. Full code shares per dimension are reported in Table~\ref{tab:5}.

\begin{table*}[tp]
\centering
\caption{Code shares within engaged data points, per CSAF dimension.}
\label{tab:5}
\footnotesize
\begin{tabular}{L{4.8cm} L{4.8cm} C{2.6cm} R{1.5cm} R{2.7cm}}
\toprule
\textbf{Dimension} & \textbf{Code} & \textbf{Valence} & \textbf{$n$} & \textbf{Code share (\%)} \\
\midrule
Principle Integrity ($n = 1{,}288$) & COMMITMENT-FIRM & Favorable & 802 & 62.3 \\
 & COMMITMENT-YIELDING & Unfavorable & 321 & 24.9 \\
 & COMMITMENT-FLEXIBLE & Unfavorable & 165 & 12.8 \\
\addlinespace
Self-Efficacy ($n = 802$) & EFFICACY-CONFIDENT & Favorable & 627 & 78.2 \\
 & EFFICACY-DOUBTFUL & Unfavorable & 120 & 15.0 \\
 & EFFICACY-EXTERNAL & Unfavorable & 55 & 6.9 \\
\addlinespace
Compliance Orientation ($n = 309$) & COMPLIANCE-COMMITTED & Favorable & 164 & 53.1 \\
 & COMPLIANCE-OBLIGATED & Unfavorable & 71 & 23.0 \\
 & COMPLIANCE-RESISTANT & Unfavorable & 60 & 19.4 \\
 & COMPLIANCE-CONSTRAINED & Unfavorable & 14 & 4.5 \\
\addlinespace
Learning Orientation ($n = 764$) & LEARNING-ADAPTIVE & Favorable & 653 & 85.5 \\
 & LEARNING-INDIFFERENT & Unfavorable & 52 & 6.8 \\
 & LEARNING-DEFENSIVE & Unfavorable & 33 & 4.3 \\
 & LEARNING-REINFORCING & Unfavorable & 26 & 3.4 \\
\addlinespace
Risk Perception ($n = 3{,}502$) & RISK-AWARE & Favorable & 2{,}725 & 77.8 \\
 & RISK-MINIMIZED & Unfavorable & 728 & 20.8 \\
 & RISK-UNCERTAIN & Unfavorable & 49 & 1.4 \\
\addlinespace
Benefit Perception ($n = 1{,}747$) & BENEFIT-POSITIVE & Favorable & 1{,}407 & 80.5 \\
 & BENEFIT-NEGATIVE & Unfavorable & 279 & 16.0 \\
 & BENEFIT-SKEPTICAL & Unfavorable & 61 & 3.5 \\
\addlinespace
Social Norms ($n = 1{,}247$) & NORMS-DISCOURAGING & Unfavorable & 865 & 69.4 \\
 & NORMS-INDIFFERENT & Unfavorable & 196 & 15.7 \\
 & NORMS-SUPPORTIVE & Favorable & 186 & 14.9 \\
\addlinespace
Personal Agency ($n = 465$) & AGENCY-ACTIVE & Favorable & 294 & 63.2 \\
 & AGENCY-PASSIVE & Unfavorable & 154 & 33.1 \\
 & AGENCY-NEGATIVE & Unfavorable & 17 & 3.7 \\
\bottomrule
\end{tabular}
\end{table*}

\begin{figure}[tb]
  \centering
  \includegraphics[width=\linewidth]{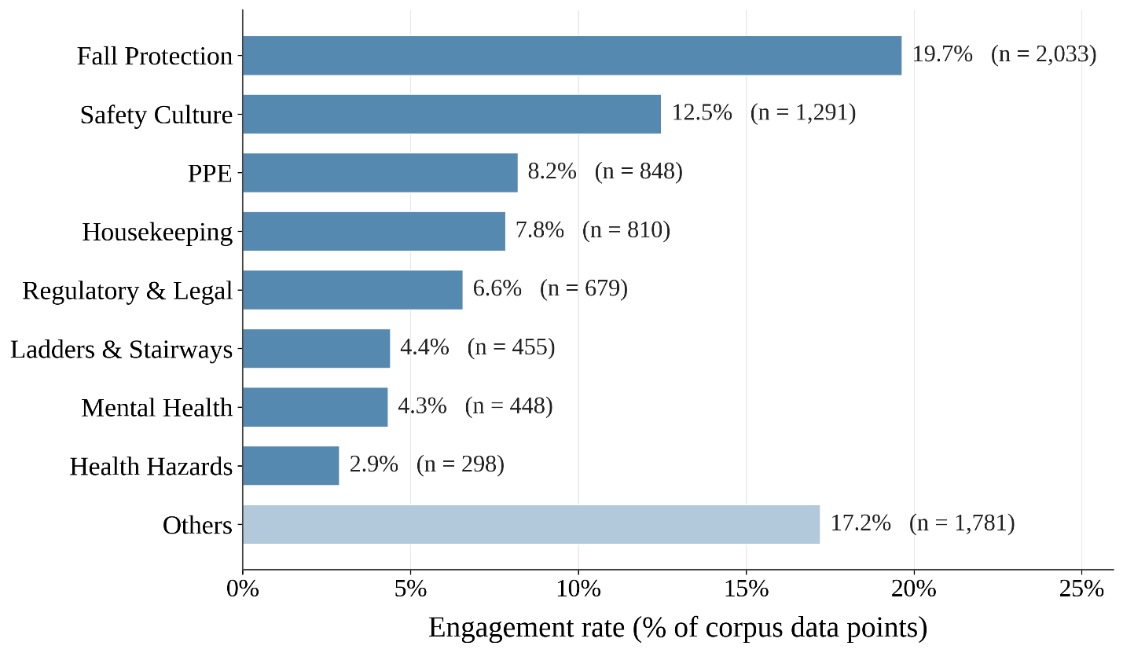}
  \caption{Engagement rate per safety topic.}
  \label{fig:12}
\end{figure}

\begin{figure}[tb]
  \centering
  \includegraphics[width=\linewidth]{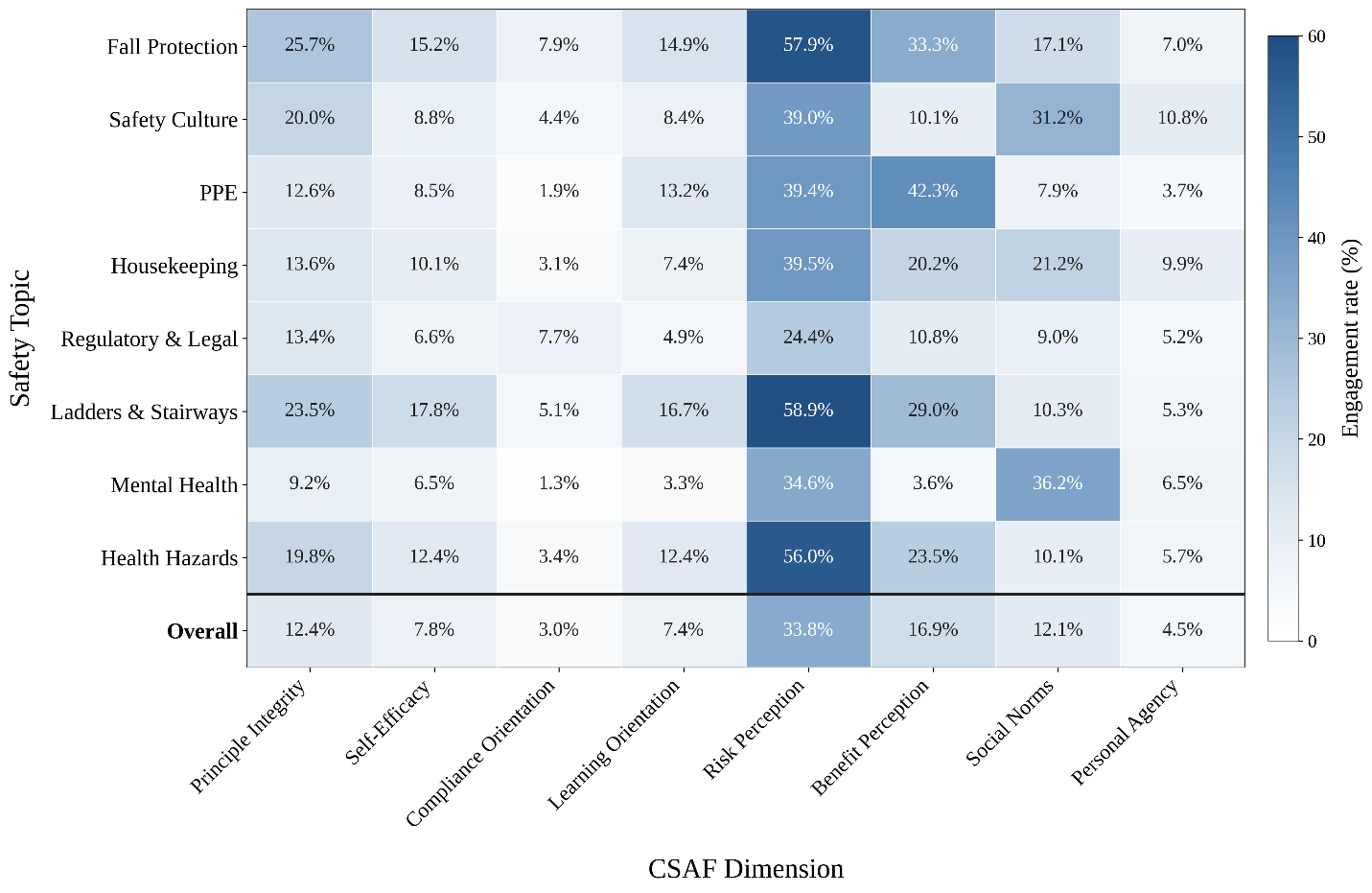}
  \caption{Engagement rate by safety topic $\times$ CSAF dimension.}
  \label{fig:13}
\end{figure}

Topic stratification extended the engagement rate and $F/U$ ratio to the safety-topic level. The eight most-engaged topics were retained for analysis (Fig.~\ref{fig:12}): Fall Protection, Safety Culture \& Attitudes, PPE, Housekeeping \& Sanitation, Regulatory \& Legal, Ladders \& Stairways, Mental Health \& Wellbeing, and Health Hazard Exposures. Engagement varied across both safety topics and CSAF dimensions (Fig.~\ref{fig:13}). Risk Perception was the most-engaged dimension in three topics: Ladders \& Stairways at 58.9\%, Fall Protection at 57.9\%, and Health Hazards at 56.0\%. Benefit Perception was the most-engaged dimension in PPE (42.3\%), and Social Norms was the most-engaged dimension in Mental Health (36.2\%). Collectively, the results show that the CSAF dimensions raised by workers shifted with the topic at hand rather than reflecting a single evaluative lens applied uniformly across topics.

\begin{figure}[tb]
  \centering
  \includegraphics[width=\linewidth]{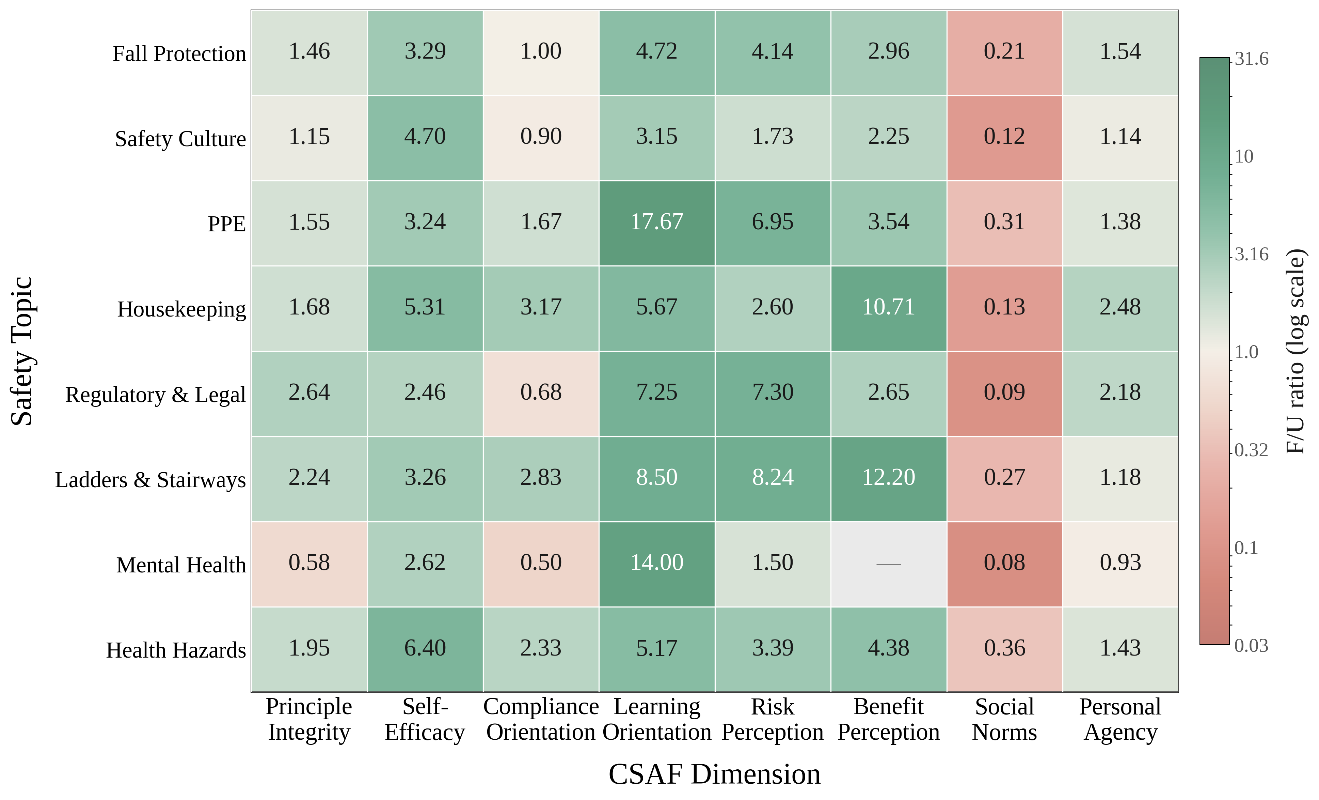}
  \caption{Favorable-to-unfavorable ($F/U$) ratio across safety topics.}
  \label{fig:14}
\end{figure}

As shown in Fig.~\ref{fig:14}, the topic-level $F/U$ ratio revealed that Social Norms skewed unfavorable across all eight topics, with $F/U$ ratios ranging from 0.08 in Mental Health to 0.36 in Health Hazards. Compliance Orientation skewed unfavorable in three topics: Mental Health (0.50), Regulatory \& Legal (0.68), and Safety Culture (0.90). Mental Health emerged as the most attitudinally complex topic, as it was the only topic where unfavorable Social Norms and Compliance Orientation converged and the only topic to additionally skew unfavorable on Principle Integrity ($F/U = 0.58$). Together, these three unfavorable dimensions suggest that workers perceived weak institutional compliance, questionable peer norms, and compromised safety principles in the context of mental health. At the same time, the same discourse exhibited a strongly favorable Learning Orientation ($F/U = 14.00$), second only to PPE (17.67), indicating that workers simultaneously expressed a pronounced openness to learning and adaptation. This internally divided attitudinal profile would likely remain undetected by unidimensional measures of safety attitudes, underscoring the analytical value of the multidimensional CSAF framework.

\subsubsection{Temporal Trends in Safety Attitudes}

\begin{figure}[tb]
  \centering
  \includegraphics[width=\linewidth]{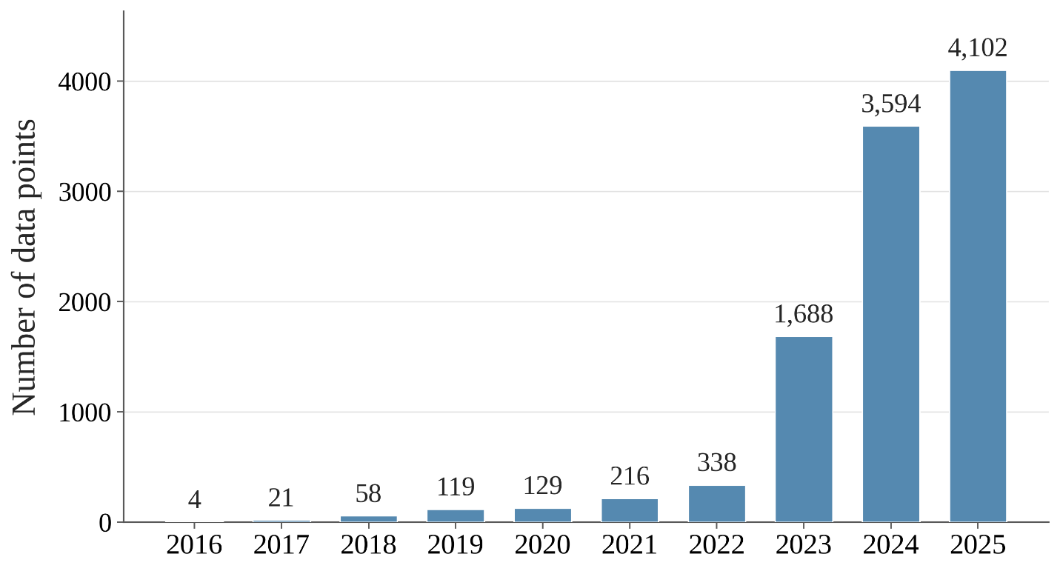}
  \caption{Number of data points in the corpus by year, 2016 to 2025.}
  \label{fig:15}
\end{figure}

Safety-related discourse in the r/Roofing community grew substantially over the study period (Fig.~\ref{fig:15}). Annual data points increased from 4 in 2016 to 4,102 in 2025, and the six years from 2020 to 2025 account for 97.3\% of the corpus. This growth establishes social media as a substantial record of safety attitudes and provides the data density required for a per-year analysis from 2020 to 2025.

\begin{figure}[tb]
  \centering
  \includegraphics[width=\linewidth]{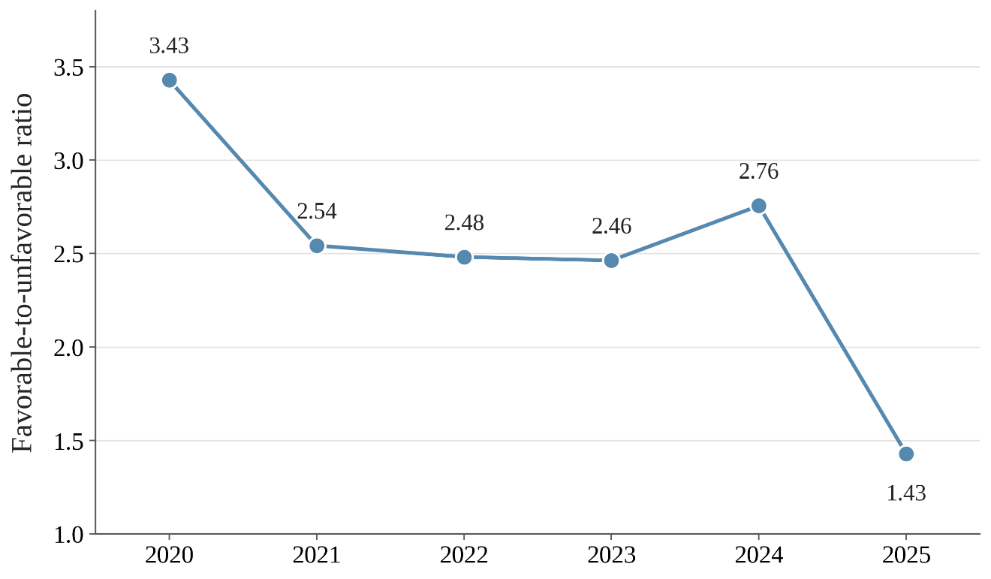}
  \caption{Favorable-to-unfavorable ($F/U$) ratio by year, 2020 to 2025.}
  \label{fig:16}
\end{figure}

The aggregate $F/U$ ratio, pooled across all eight CSAF dimensions, declined over this period (Fig.~\ref{fig:16}). The ratio measured 3.43 in 2020, remained between 2.46 and 2.76 from 2021 to 2024, and fell to 1.43 in 2025, the lowest value in the six-year series. Discourse volume increased over the same interval; the 2025 decline therefore reflects a higher proportion of unfavorable evaluations rather than a higher count alone.

\begin{figure}[tb]
  \centering
  \includegraphics[width=\linewidth]{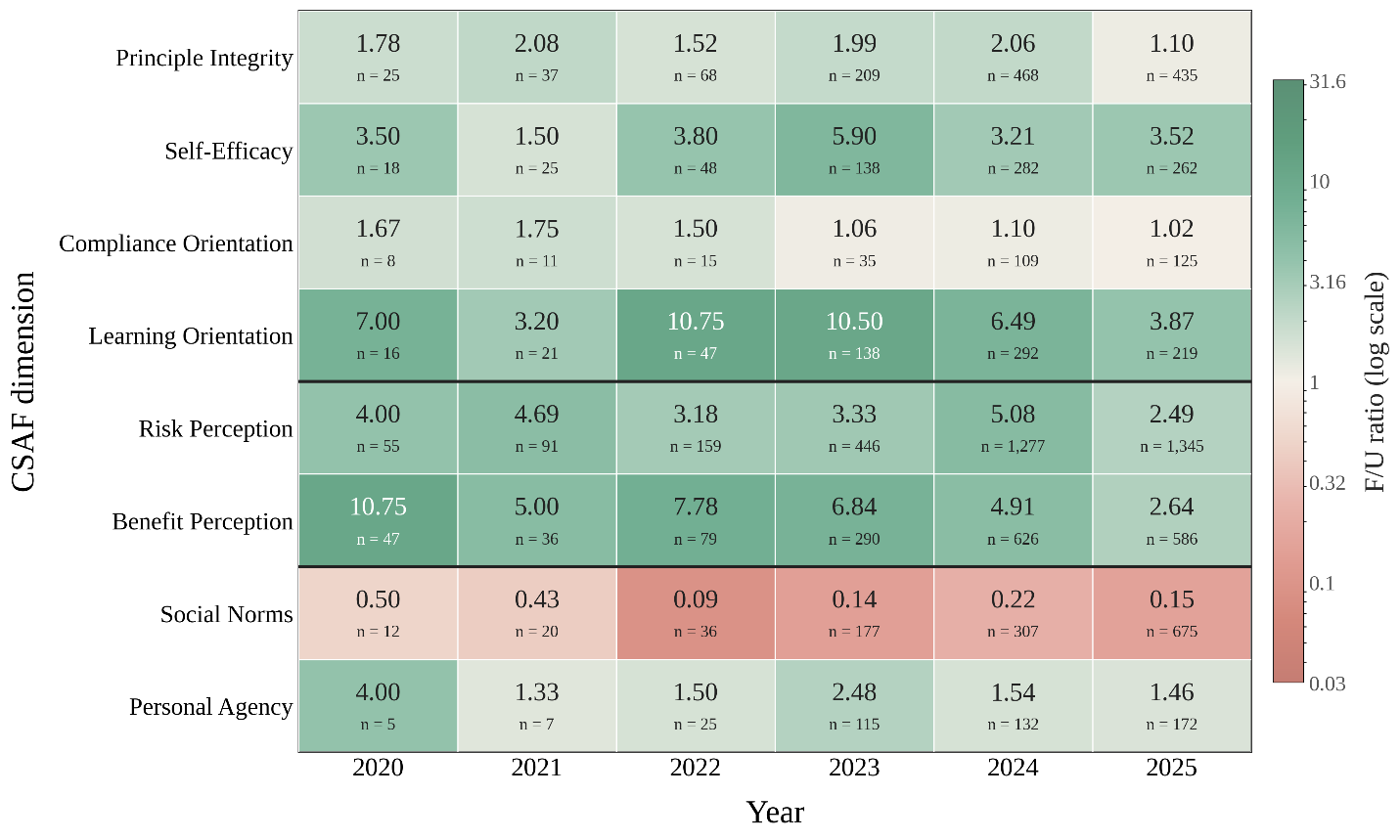}
  \caption{Favorable-to-unfavorable ($F/U$) ratio by CSAF dimension and year, 2020 to 2025.}
  \label{fig:17}
\end{figure}

The $F/U$ ratio resolved by dimension identifies four dimensions that declined from 2024 to 2025 (Fig.~\ref{fig:17}): Benefit Perception from 4.91 to 2.64, Learning Orientation from 6.49 to 3.87, Risk Perception from 5.08 to 2.49, and Principle Integrity from 2.06 to 1.10. These four dimensions span attitudinal evaluations of hazards, protective measures, safety priority, and openness to learning, indicating that the decline reached across multiple components of safety attitude rather than concentrating on one type of evaluation. The remaining three dimensions did not decline over the same interval: Self-Efficacy increased marginally, and Compliance Orientation and Personal Agency each changed by less than 0.10.

Social Norms exhibited a distinct pattern (Fig.~\ref{fig:17}). The Social Norms $F/U$ ratio remained below 1.0 in each of the six years from 2020 to 2025, the only dimension to remain unfavorable throughout. The same imbalance was reported at the dimensional and topic levels (Section~5.3.1), where Social Norms was the only dimension that skewed unfavorable overall and skewed unfavorable across all eight safety topics. The unfavorable evaluation of the peer environment therefore held across dimensions, safety topics, and years. The 2025 change in Social Norms was one of frequency rather than valence: the data points engaging the dimension increased from 307 in 2024 to 675 in 2025, and the dimension share of all engaged evaluations increased from 8.8\% to 17.7\%. Because Social Norms is unfavorable in every year, this increase in its discourse share reduced the aggregate $F/U$ ratio independently of any change within the dimension.

\begin{figure}[tb]
  \centering
  \includegraphics[width=\linewidth]{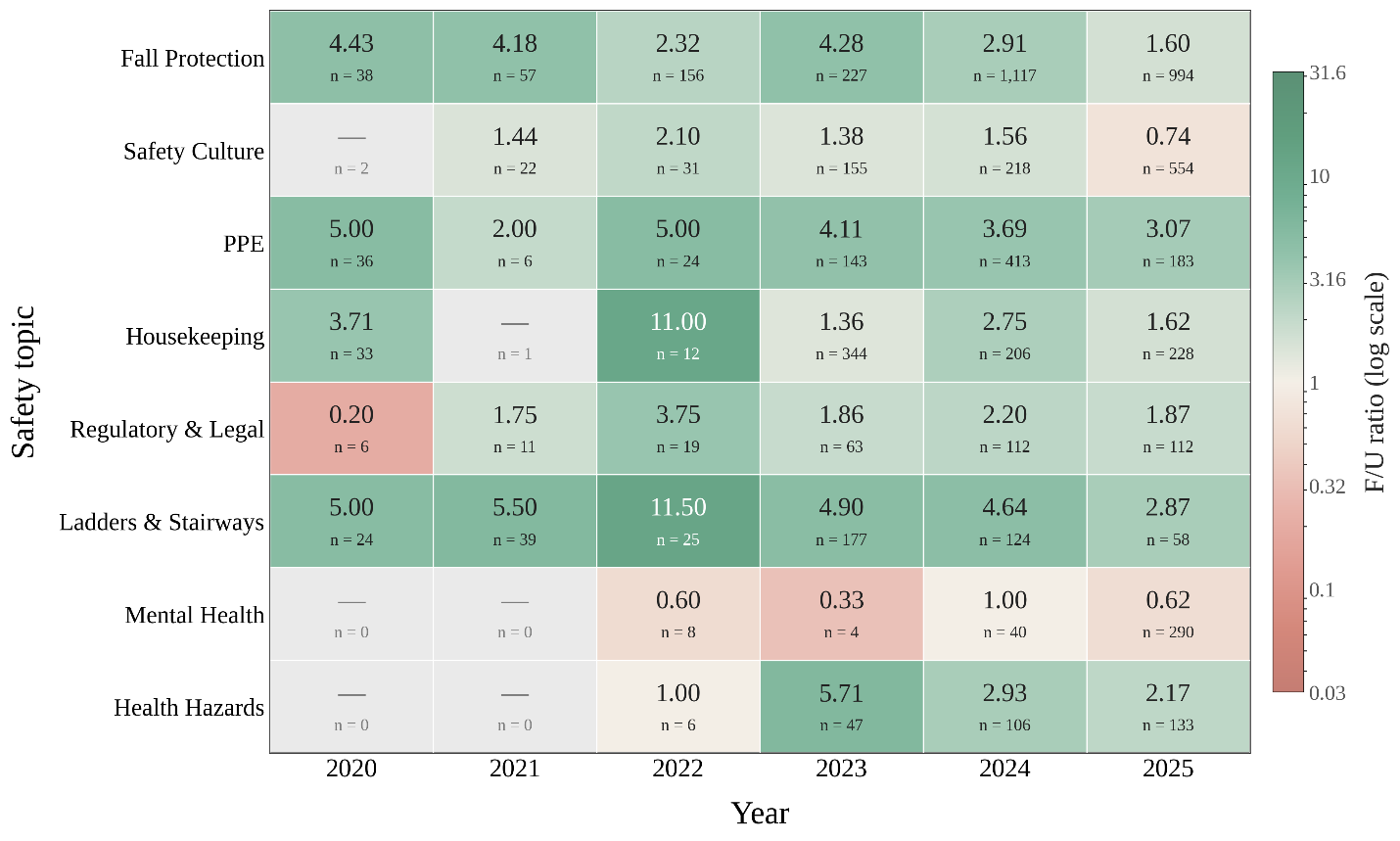}
  \caption{Favorable-to-unfavorable ($F/U$) ratio by safety topic and year, 2020 to 2025. Cells report the $F/U$ ratio and the corresponding data-point count.}
  \label{fig:18}
\end{figure}

The decline was also present at the topic level (Fig.~\ref{fig:18}), which reports the $F/U$ ratio for the eight most-engaged topics by year. The ratio declined in all eight topics from 2024 to 2025; Fall Protection, the most-engaged topic, declined from 2.65 to 1.45. Two topics fell below 1.0, the threshold at which unfavorable evaluations outnumber favorable evaluations: Safety Culture \& Attitudes reached 0.78, and Mental Health \& Wellbeing reached 0.66 as its annual volume increased from 81 to 346 data points. Mental Health \& Wellbeing also recorded the most unfavorable Social Norms balance of any topic in the dimensional results (Section~5.3.1). The ratio declined within every topic; the decline is therefore not attributable to a change in which topics drew discussion.

\subsubsection{Qualitative Analysis of Unfavorable Attitudes}

CSAF was used to identify and filter data points expressing unfavorable safety attitudes within a single topic, after which thematic analysis surfaced the underlying reasoning. This filter-and-theme procedure demonstrates CSAF's utility in informing targeted interventions that address specific psychological barriers. In this case study, Fall Protection was selected as the focal topic because it ranked as the most-engaged topic in the corpus (19.6\%). Within Fall Protection, the analysis focused on Risk Perception and Benefit Perception, the two most-engaged CSAF dimensions within Fall Protection (57.9\% and 33.3\%, respectively). Analysis of the classified data points yielded seven themes (Table~\ref{tab:6}).

\begin{table*}[tp]
\centering
\caption{Themes of unfavorable Risk Perception and Benefit Perception within Fall Protection.}
\label{tab:6}
\footnotesize
\begin{tabular}{L{3.4cm} L{6.2cm} C{1.3cm} L{5.0cm}}
\toprule
\textbf{Theme} & \textbf{Definition} & \textbf{Freq.} & \textbf{Exemplar Quote} \\
\midrule
Risk -- Complacency from Uneventful Experience & Extended periods without a fall are interpreted as evidence that fall risk is negligible, treating accumulated luck as proof of safety. & 103 & \textit{``Our old crew never tied off. We never had any accidents''} \\
\addlinespace
Benefit -- Skill as Substitute for Fall Protection & Personal balance, grip, and judgment are treated as sufficient safeguards against falls, substituting personal attributes for mechanical fall protection. & 22 & \textit{``We dont wear em cuz we have the experience and know how to maintain grip.''} \\
\addlinespace
Risk -- Underestimation of Low-Height and Low-Pitch Fall Risk & Low-pitch and low-height roofs are reclassified as ``ground level'' work, with the energy of a fall assumed insufficient to cause serious injury. & 20 & \textit{``That is a very very low pitch roof. Harnesses would be a joke on that roof.''} \\
\addlinespace
Benefit -- Fall Protection as a Source of New Hazards & Harnesses, lanyards, and ropes are perceived as sources of tripping, snagging, entanglement, and suspension trauma that increase rather than reduce fall risk. & 17 & \textit{``a harness is more likely to make you hurt yourself tripping than keep you from falling''} \\
\addlinespace
Risk -- Misjudgment of Surface and Environmental Conditions & The structural soundness of roof surfaces and the hazard profile of weather and environmental conditions are judged through visual checks rather than structural assessment. & 14 & \textit{``Give it a kick or two, then you'll be sure''} \\
\addlinespace
Benefit -- Fall Protection as Disproportionate & Fall protection is dismissed as requiring setup time and effort disproportionate to the duration and risk of the task. & 9 & \textit{``Takes too long and it's more of a hassle than it is a help''} \\
\addlinespace
Benefit -- Fall Protection as Unreliable & Fall protection is perceived as functionally unreliable, with improper anchorage, inadequate arrest distance, and cosmetic compliance treated as evidence the system fails to deliver real protection. & 6 & \textit{``My boss only breaks out the harnesses for show.''} \\
\bottomrule
\end{tabular}
\end{table*}

Three themes captured how workers reasoned about risk in fall protection contexts. First, personal experience operated as a substitute for population-level evidence, with incident-free personal history frequently invoked as proof that the hazard did not apply ($n = 103$). Second, workers restricted the relevance of fall protection to extreme heights and steep pitches ($n = 20$), a misjudgment that departs from regulatory thresholds. Third, workers judged roof surfaces and environmental conditions to be safe on the basis of visual familiarity ($n = 14$), with familiar conditions routinely perceived as less risky, such as skylights, coated panels, weathered decking, and snow cover. The remaining four themes captured how workers evaluated the benefits of safety controls. First, workers, particularly those self-reported with extensive experience, framed fall protection as redundant given personal skill ($n = 22$), reserving harnesses for the inexperienced. Workers also framed harnesses, lanyards, and ropes as injurious rather than protective ($n = 17$), citing tripping, snagging, suspension trauma, and entanglement as common concerns. Workers further argued that setup time could exceed the protective value of the equipment for short-duration tasks ($n = 9$). Finally, workers framed fall protection as functionally unreliable in field conditions ($n = 6$), pointing to inadequate arrest distance at low heights as a representative failure mode.

Across the seven themes, the findings point to a two-pronged intervention strategy for overcoming unfavorable attitudes toward Risk Perception and Benefit Perception in fall protection. The first prong requires reshaping how workers reason about low-probability, high-consequence events in order to dismantle overconfidence and complacency, while the second requires addressing the design, reliability, and usability of the equipment itself. By filtering discourse to a single topic and valence before thematic analysis, the framework isolates the specific psychological barriers operating in a workforce and allows interventions to be matched to the mechanisms that sustain them.

\section{Discussion}

This study developed and validated the Construction Safety Attitude Framework (CSAF), a multidimensional classification system that measures construction worker safety attitudes from naturalistic discourse through LLM-based classification. The framework demonstrated strong performance across multiple validation layers, including human-coder reliability (Krippendorff's $\alpha = 0.85$, 0.86, 0.81 across development, validation, and transfer), LLM--human agreement (Cohen's $\kappa = 0.90$ within-domain, 0.89 on transfer), classification stability (97\% code-level agreement), and construct distinctness (zero of 254 defined code pairs exceeded the psychometric redundancy threshold). Applied to a case-study corpus of 10,346 posts and comments, CSAF revealed attitudinal patterns at the dimensional, topic, and temporal levels. These results indicate that carefully structured LLM-based classification can reliably operationalize multidimensional psychological frameworks for workforce-scale safety attitude measurement. By making safety attitudes legible in workers' own language, the framework opens a route for safety science to draw on worker voice at the scale that engineering and management data have long supported.

The contribution of CSAF extends beyond its validation. The framework's eight dimensions cover a range of evaluative targets that workers raise in safety discourse, spanning the physical work environment, the social and organizational context, and workers' own capability, responsibility, and openness to learning. CSAF integrates constructs from Value Theory \citep{schwartz1992}, the Theory of Planned Behavior \citep{ajzen1991}, Tripartite Attitude Theory \citep{rosenberg1960}, Social Cognitive Theory \citep{bandura1986}, the Health Belief Model \citep{rosenstock1974}, Social Identity Theory \citep{tajfel2004}, and Dual Attitude Theory \citep{wilson2000} into a single instrument, which can measure constructs developed in separate traditions jointly on the same utterances without losing statistical separability.

Applying CSAF to roofing-trade discourse demonstrated the framework's practical value through three findings about how safety attitudes are structured. The first concerns code co-occurrence. Unfavorable codes co-occurred more strongly with one another than favorable codes co-occurred among themselves, consistent with negativity bias under which negative evaluations carry greater psychological weight than positive evaluations of comparable magnitude \citep{baumeister2001}, and with attitude-network research showing that evaluative components are interconnected and same-valence components tend to co-activate \citep{dalege2016}. The practical implication is that unfavorable attitudes appear jointly. Interventions targeting one unfavorable dimension in isolation are likely to be undermined by adjacent unfavorable dimensions left unaddressed.

The second finding concerns the peer environment. Social Norms was the only dimension that skewed unfavorable when engaged ($F/U = 0.18$), with the imbalance holding across all eight safety topics and all six years of dense coverage. The Social Norms imbalance admits two readings the data does not decisively distinguish. The peer environment on roofing sites may genuinely discourage safety engagement, consistent with occupational identity research showing that trades define competence through risk tolerance \citep{iacuone2005, stergioukita2015}, or pseudonymous discourse may surface frustration that face-to-face settings keep private \citep{kapoor2017}. Both readings interpret expressed evaluations through a cognitive-evaluative lens; the affective register of peer-environment hostility falls outside CSAF's current operationalization. Either reading converges on the same practical signal: greater attention to the social and peer context is warranted in safety intervention design.

The third finding concerns change over time. The aggregate $F/U$ ratio declined from 2.46--2.76 across 2021--2024 to 1.43 in 2025, with the decline registering across four of the eight dimensions. \citet{xu2023} demonstrated that group-level factors drive the severity of safety-attitude loss under disruption while individual-level factors govern recovery. CSAF's dimension-level measurement provides a methodology suited to tracking such dynamics in ongoing time-series alongside established survey-based approaches.

The implications of CSAF for safety management lie in its thematic analysis capability, which decomposes attitudes into the reasoning patterns that drive them and links each to a distinct intervention type. The fall-protection analysis illustrates this: unfavorable codes within two dimensions resolved into reasoning patterns that pointed to interventions as varied as training content, equipment design, and procurement planning, each addressing a different actor in the safety management chain. Wherever CSAF is applied, the unfavorable codes within a dimension can be filtered to surface the specific patterns driving them, and those patterns can be mapped to the actors and intervention types best placed to address them. The framework therefore equips construction safety research, training, and policy with a diagnostic that locates where attitudinal barriers concentrate and what intervention each pattern requires. Cross-domain transfer without retraining indicates that the same diagnostic instrument can extend to other construction subtrades and discourse sources.

\section{Limitations and Future Research}

The study operates within four scope conditions that shape interpretation and point to priorities for future research. First, participation in a trade subreddit is self-selected, and the posting population may differ from the broader construction workforce. Future work should triangulate CSAF profiles against survey and implicit-association measures within the same workforce, and apply the framework to discourse sources that reach populations underrepresented on Reddit. Second, CSAF measures attitudes primarily through a cognitive-evaluative lens, without accounting for affective registers such as fear, frustration, and pride. Extending the framework to multimodal worker discourse would enable analysis through a combined cognitive-affective lens. Third, CSAF classifies the expressed attitude but does not identify the driver behind it. Pairing CSAF with driver-elicitation methods would convert it from a diagnostic of what workers express into a tool for explaining why specific patterns concentrate on specific topics and trades. Fourth, the present study is observational; it characterizes attitudes but does not test how they respond to intervention. Longitudinal designs that pair CSAF with intervention trials could track how attitudes shift in response to training, equipment design, and regulatory change.

\section{Conclusion}

CSAF establishes a multidimensional instrument for measuring construction worker safety attitudes from naturalistic discourse. Operationalized as a structured codebook embedded in LLM classification prompts, the framework achieves substantial agreement with expert-human consensus on both within-domain validation and cross-domain transfer, surfaces reasoning patterns within unfavorable attitudes, and links each pattern to a distinct intervention type spanning training, equipment design, and procurement planning. The same instrument can extend to other construction subtrades and discourse sources, bringing workforce attitudinal evidence within reach of construction safety research. By making safety attitudes legible in workers' own language, CSAF equips the field to draw on worker voice at the scale that engineering and management data have long supported.

\section*{Declaration of Generative AI and AI-assisted technologies in the writing process}

During the preparation of this work the authors used ChatGPT in order to improve language. After using this tool, the authors reviewed and edited the content as needed and take full responsibility for the content of the published article.

\section*{Declaration of competing interest}

The authors declare that they have no known competing financial interests or personal relationships that could have appeared to influence the work reported in this paper.

\section*{Acknowledgments}

The authors thank the Roofing Alliance for industry expertise that grounded the practical interpretation of findings in roofing-trade realities. The authors also thank Reddit, Inc.\ for access to the r/Construction and r/Roofing data through the Reddit for Researchers (RFR) program, which enabled the naturalistic, large-scale discourse analysis central to this work.

\bibliographystyle{elsarticle-harv}
\bibliography{references}

@article{ahmadi2024,
  author  = {Ahmadi, Ehsan and Muley, Suraj and Wang, Chao},
  title   = {Automatic Construction Accident Report Analysis Using Large Language Models ({LLMs})},
  journal = {Journal of Intelligent Construction},
  year    = {2024},
  volume  = {3},
  pages   = {1--10},
  doi     = {10.26599/JIC.2024.9180039}
}

@article{ajzen1991,
  author  = {Ajzen, Icek},
  title   = {The theory of planned behavior},
  journal = {Organizational Behavior and Human Decision Processes},
  year    = {1991},
  volume  = {50},
  pages   = {179--211},
  doi     = {10.1016/0749-5978(91)90020-T}
}

@book{bandura1986,
  author    = {Bandura, Albert},
  title     = {Social Foundations of Thought and Action: A Social Cognitive Theory},
  publisher = {Prentice-Hall},
  address   = {Englewood Cliffs, NJ},
  year      = {1986}
}

@inproceedings{bao2023,
  author    = {Bao, Guangsheng and Zhao, Yanbin and Teng, Zhiyang and Yang, Linyi and Zhang, Yue},
  title     = {Fast-{DetectGPT}: Efficient Zero-Shot Detection of Machine-Generated Text via Conditional Probability Curvature},
  booktitle = {Proceedings of the 12th International Conference on Learning Representations (ICLR)},
  year      = {2023}
}

@article{basahel2021,
  author  = {Basahel, Abdulrahman M.},
  title   = {Safety Leadership, Safety Attitudes, Safety Knowledge and Motivation toward Safety-Related Behaviors in Electrical Substation Construction Projects},
  journal = {International Journal of Environmental Research and Public Health},
  year    = {2021},
  volume  = {18},
  pages   = {4196},
  doi     = {10.3390/IJERPH18084196}
}

@article{baumeister2001,
  author  = {Baumeister, Roy F. and Bratslavsky, Ellen and Finkenauer, Catrin and Vohs, Kathleen D.},
  title   = {Bad Is Stronger Than Good},
  journal = {Review of General Psychology},
  year    = {2001},
  volume  = {5},
  pages   = {323--370},
  doi     = {10.1037/1089-2680.5.4.323}
}

@inproceedings{brin1997,
  author    = {Brin, Sergey and Motwani, Rajeev and Ullman, Jeffrey D. and Tsur, Shalom},
  title     = {Dynamic itemset counting and implication rules for market basket data},
  booktitle = {Proceedings of the ACM SIGMOD International Conference on Management of Data},
  year      = {1997},
  pages     = {255--264},
  doi       = {10.1145/253260.253325}
}

@misc{bls2024,
  author = {{Bureau of Labor Statistics}},
  title  = {Census of Fatal Occupational Injuries Summary, 2023},
  year   = {2024},
  note   = {Economic News Release},
  howpublished = {\url{https://www.bls.gov/news.release/cfoi.nr0.htm}}
}

@article{burns2013,
  author  = {Burns, Calvin and Conchie, Stacey},
  title   = {Risk information source preferences in construction workers},
  journal = {Employee Relations},
  year    = {2013},
  volume  = {36},
  pages   = {70--81},
  doi     = {10.1108/ER-06-2013-0060}
}

@article{campbell2013,
  author  = {Campbell, John L. and Quincy, Charles and Osserman, Jordan and Pedersen, Ove K.},
  title   = {Coding In-depth Semistructured Interviews: Problems of Unitization and Intercoder Reliability and Agreement},
  journal = {Sociological Methods \& Research},
  year    = {2013},
  volume  = {42},
  pages   = {294--320},
  doi     = {10.1177/0049124113500475}
}

@article{cavazza2009,
  author  = {Cavazza, Nicoletta and Serpe, Alessandra},
  title   = {Effects of safety climate on safety norm violations: exploring the mediating role of attitudinal ambivalence toward personal protective equipment},
  journal = {Journal of Safety Research},
  year    = {2009},
  volume  = {40},
  pages   = {277--283},
  doi     = {10.1016/J.JSR.2009.06.002}
}

@article{clarke2017,
  author  = {Clarke, Victoria and Braun, Virginia},
  title   = {Thematic analysis},
  journal = {The Journal of Positive Psychology},
  year    = {2017},
  volume  = {12},
  pages   = {297--298},
  doi     = {10.1080/17439760.2016.1262613}
}

@article{cohen1960,
  author  = {Cohen, Jacob},
  title   = {A Coefficient of Agreement for Nominal Scales},
  journal = {Educational and Psychological Measurement},
  year    = {1960},
  volume  = {20},
  pages   = {37--46},
  doi     = {10.1177/001316446002000104}
}

@article{dalege2016,
  author  = {Dalege, Jonas and Borsboom, Denny and van Harreveld, Frenk and van den Berg, Helma and Conner, Mark and van der Maas, Han L. J.},
  title   = {Toward a formalized account of attitudes: The Causal Attitude Network ({CAN}) Model},
  journal = {Psychological Review},
  year    = {2016},
  volume  = {123},
  pages   = {2--22},
  doi     = {10.1037/A0039802}
}

@incollection{dane2019,
  author    = {Dane, Francis C.},
  title     = {Survey Methods, Naturalistic Observations, and Case-Studies},
  booktitle = {Companion Encyclopedia of Psychology},
  year      = {2019},
  pages     = {1142--1155},
  doi       = {10.4324/9781315542072-30}
}

@book{daniel2023,
  author    = {Riffe, Daniel and Lacy, Stephen and Watson, Brendan and Lovejoy, Jennette},
  title     = {Analyzing Media Messages: Using Quantitative Content Analysis in Research},
  edition   = {5},
  publisher = {Routledge},
  year      = {2023},
  doi       = {10.4324/9781003288428}
}

@book{eagly1993,
  author    = {Eagly, Alice H. and Chaiken, Shelly},
  title     = {The Psychology of Attitudes},
  publisher = {Harcourt Brace Jovanovich},
  year      = {1993}
}

@article{fang2016,
  author  = {Fang, Dongping and Zhao, Chen and Zhang, Mingyuan},
  title   = {A Cognitive Model of Construction Workers' Unsafe Behaviors},
  journal = {Journal of Construction Engineering and Management},
  year    = {2016},
  volume  = {142},
  pages   = {04016039},
  doi     = {10.1061/(ASCE)CO.1943-7862.0001118}
}

@inproceedings{gharibi2016,
  author    = {Gharibi, Vahid and Mortazavi, Seyed Bagher and Jafari, Ahmad Jonidi and Malakouti, Jamshid and Abadi, Mohammad Bagher Heydari},
  title     = {The relationship between workers' attitude towards safety and occupational accidents experience},
  booktitle = {International Journal of Occupational Hygiene},
  year      = {2016}
}

@misc{grandini2020,
  author = {Grandini, Margherita and Bagli, Enrico and Visani, Giorgio},
  title  = {Metrics for Multi-Class Classification: an Overview},
  year   = {2020},
  note   = {arXiv preprint}
}

@book{grimmer2022,
  author    = {Grimmer, Justin and Roberts, Margaret E. and Stewart, Brandon M.},
  title     = {Text as Data: A New Framework for Machine Learning and the Social Sciences},
  publisher = {Princeton University Press},
  year      = {2022}
}

@article{guest2006,
  author  = {Guest, Greg and Bunce, Arwen and Johnson, Laura},
  title   = {How Many Interviews Are Enough? An Experiment with Data Saturation and Variability},
  journal = {Field Methods},
  year    = {2006},
  volume  = {18},
  pages   = {59--82},
  doi     = {10.1177/1525822X05279903}
}

@article{hahsler2005,
  author  = {Hahsler, Michael and Gr{\"u}n, Bettina and Hornik, Kurt},
  title   = {arules -- A Computational Environment for Mining Association Rules and Frequent Item Sets},
  journal = {Journal of Statistical Software},
  year    = {2005},
  volume  = {14},
  pages   = {1--25},
  doi     = {10.18637/JSS.V014.I15}
}

@article{hahsler2017,
  author  = {Hahsler, Michael and Karpienko, Radoslaw},
  title   = {arulesViz: Interactive Visualization of Association Rules with {R}},
  journal = {The R Journal},
  year    = {2017}
}

@book{hair2019,
  author    = {Hair, Joseph F. and Babin, Barry J. and Black, William C. and Anderson, Rolph E.},
  title     = {Multivariate Data Analysis},
  publisher = {Cengage},
  year      = {2019}
}

@article{hennink2017,
  author  = {Hennink, Monique M. and Kaiser, Bonnie N. and Marconi, Vincent C.},
  title   = {Code Saturation Versus Meaning Saturation: How Many Interviews Are Enough?},
  journal = {Qualitative Health Research},
  year    = {2017},
  volume  = {27},
  pages   = {591--608},
  doi     = {10.1177/1049732316665344}
}

@article{hsieh2005,
  author  = {Hsieh, Hsiu-Fang and Shannon, Sarah E.},
  title   = {Three approaches to qualitative content analysis},
  journal = {Qualitative Health Research},
  year    = {2005},
  volume  = {15},
  pages   = {1277--1288},
  doi     = {10.1177/1049732305276687}
}

@article{hu2023,
  author  = {Hu, Zhenyu and Chan, Wee Tiong and Hu, Hao and Xu, Feng},
  title   = {Cognitive Factors Underlying Unsafe Behaviors of Construction Workers as a Tool in Safety Management: A Review},
  journal = {Journal of Construction Engineering and Management},
  year    = {2023},
  volume  = {149},
  pages   = {03123001},
  doi     = {10.1061/JCEMD4.COENG-11820}
}

@article{hung2011,
  author  = {Hung, Yu-Hsiu and Smith-Jackson, Tonya and Winchester, Woodrow},
  title   = {Use of attitude congruence to identify safety interventions for small residential builders},
  journal = {Construction Management and Economics},
  year    = {2011},
  volume  = {29},
  pages   = {113--130},
  doi     = {10.1080/01446193.2010.521758}
}

@article{iacuone2005,
  author  = {Iacuone, David},
  title   = {``Real Men are Tough Guys'': Hegemonic Masculinity and Safety in the Construction Industry},
  journal = {The Journal of Men's Studies},
  year    = {2005},
  volume  = {13},
  pages   = {247--266},
  doi     = {10.3149/JMS.1302.247}
}

@article{jhaver2019,
  author  = {Jhaver, Shagun and Birman, Iris and Gilbert, Eric and Bruckman, Amy},
  title   = {Human-machine collaboration for content regulation: The case of reddit automoderator},
  journal = {ACM Transactions on Computer-Human Interaction},
  year    = {2019},
  volume  = {26},
  doi     = {10.1145/3338243}
}

@article{johari2020,
  author  = {Johari, Saumyendu and Jha, Kumar Neeraj},
  title   = {Interrelationship among Belief, Intention, Attitude, Behavior, and Performance of Construction Workers},
  journal = {Journal of Management in Engineering},
  year    = {2020},
  volume  = {36},
  pages   = {04020081},
  doi     = {10.1061/(ASCE)ME.1943-5479.0000851}
}

@misc{jung2025,
  author = {Jung, Minseok and Panizo, Carlos F. and Dugan, Liam and Yu, Rui and Fung, Yi and Chen, Pin-Yu and Liang, Paul Pu},
  title  = {Group-Adaptive Threshold Optimization for Robust AI-Generated Text Detection},
  year   = {2025},
  note   = {arXiv preprint}
}

@article{kao2019,
  author  = {Kao, Kuo-Yang and Spitzmueller, Christiane and Cigularov, Konstantin and Thomas, Candice L.},
  title   = {Linking safety knowledge to safety behaviours: a moderated mediation of supervisor and worker safety attitudes},
  journal = {European Journal of Work and Organizational Psychology},
  year    = {2019},
  volume  = {28},
  pages   = {206--220},
  doi     = {10.1080/1359432X.2019.1567492}
}

@article{kapoor2017,
  author  = {Kapoor, Kawaljeet Kaur and Tamilmani, Kuttimani and Rana, Nripendra P. and Patil, Pushp and Dwivedi, Yogesh K. and Nerur, Sridhar},
  title   = {Advances in Social Media Research: Past, Present and Future},
  journal = {Information Systems Frontiers},
  year    = {2017},
  volume  = {20},
  pages   = {531--558},
  doi     = {10.1007/S10796-017-9810-Y}
}

@inproceedings{kashmiri2020,
  author    = {Kashmiri, Danish and Taherpour, Fereshteh and Namian, Mostafa and Ghiasvand, Elnaz},
  title     = {Role of safety attitude: Impact on hazard recognition and safety risk perception},
  booktitle = {Construction Research Congress 2020: Safety, Workforce, and Education},
  year      = {2020},
  pages     = {583--590},
  doi       = {10.1061/9780784482872.063}
}

@book{krippendorff2018,
  author    = {Krippendorff, Klaus},
  title     = {Content Analysis: An Introduction to Its Methodology},
  publisher = {SAGE},
  year      = {2018}
}

@article{lacy1996,
  author  = {Lacy, Stephen and Riffe, Daniel},
  title   = {Sampling error and selecting intercoder reliability samples for nominal content categories},
  journal = {Journalism \& Mass Communication Quarterly},
  year    = {1996},
  volume  = {73},
  pages   = {963--973},
  doi     = {10.1177/107769909607300414}
}

@article{lacy2015,
  author  = {Lacy, Stephen and Watson, Brendan R. and Riffe, Daniel and Lovejoy, Jennette},
  title   = {Issues and best practices in content analysis},
  journal = {Journalism \& Mass Communication Quarterly},
  year    = {2015},
  volume  = {92},
  pages   = {791--811},
  doi     = {10.1177/1077699015607338}
}

@article{landis1977,
  author  = {Landis, J. Richard and Koch, Gary G.},
  title   = {The Measurement of Observer Agreement for Categorical Data},
  journal = {Biometrics},
  year    = {1977},
  volume  = {33},
  pages   = {159--174},
  doi     = {10.2307/2529310}
}

@article{langford2000,
  author  = {Langford, David and Rowlinson, Steve and Sawacha, Edwin},
  title   = {Safety behaviour and safety management: its influence on the attitudes of workers in the {UK} construction industry},
  journal = {Engineering, Construction and Architectural Management},
  year    = {2000},
  volume  = {7},
  pages   = {133--140},
  doi     = {10.1108/EB021138}
}

@article{legishion2024,
  author  = {Legishion, Joseph J. and Wachira, Isabella N. and K'Akumu, Owiti A.},
  title   = {Attitudes of Workers Toward Safety and Health Compliance in Construction Sites in Nairobi},
  journal = {Africa Habitat Review},
  year    = {2024},
  volume  = {19},
  pages   = {2910--2927}
}

@article{lewis2013,
  author  = {Lewis, Seth C. and Zamith, Rodrigo and Hermida, Alfred},
  title   = {Content Analysis in an Era of Big Data: A Hybrid Approach to Computational and Manual Methods},
  journal = {Journal of Broadcasting \& Electronic Media},
  year    = {2013},
  volume  = {57},
  pages   = {34--52},
  doi     = {10.1080/08838151.2012.761702}
}

@inproceedings{lloyd2025,
  author    = {Lloyd, Travis and Gosciak, Joseph and Nguyen, Tung and Naaman, Mor},
  title     = {AI Rules? Characterizing Reddit Community Policies Towards AI-Generated Content},
  booktitle = {Proceedings of the CHI Conference on Human Factors in Computing Systems},
  year      = {2025},
  doi       = {10.1145/3706598.3713292}
}

@article{loosemore2019,
  author  = {Loosemore, Martin and Malouf, Nataly},
  title   = {Safety training and positive safety attitude formation in the Australian construction industry},
  journal = {Safety Science},
  year    = {2019},
  volume  = {113},
  pages   = {233--243},
  doi     = {10.1016/J.SSCI.2018.11.029}
}

@inproceedings{mccabe2005,
  author    = {McCabe, Brenda and Karahalios, Daniela and Loughlin, Catherine},
  title     = {Attitudes in Construction Safety},
  booktitle = {Construction Research Congress 2005: Broadening Perspectives},
  year      = {2005},
  pages     = {1--9},
  doi       = {10.1061/40754(183)47}
}

@article{messick1995,
  author  = {Messick, Samuel},
  title   = {Validity of psychological assessment: Validation of inferences from persons' responses and performances as scientific inquiry into score meaning},
  journal = {American Psychologist},
  year    = {1995},
  volume  = {50},
  pages   = {741--749},
  doi     = {10.1037/0003-066X.50.9.741}
}

@article{miles2015,
  author  = {Miles, Matthew B. and Huberman, A. Michael and Salda{\~n}a, Johnny},
  title   = {Qualitative Data Analysis: A Methods Sourcebook},
  journal = {Technical Communication Quarterly},
  year    = {2015},
  volume  = {24},
  pages   = {109--112},
  doi     = {10.1080/10572252.2015.975966}
}

@article{newaz2024,
  author  = {Newaz, Mohammad Tanvi and Ershadi, Mahmoud and Jefferies, Marcus and Davis, Peter},
  title   = {A critical review of the feasibility of emerging technologies for improving safety behavior on construction sites},
  journal = {Journal of Safety Research},
  year    = {2024},
  volume  = {89},
  pages   = {269--287},
  doi     = {10.1016/J.JSR.2024.04.006}
}

@article{pandit2018,
  author  = {Pandit, Bhavana and Albert, Alex and Patil, Yashwardhan and Al-Bayati, Ahmed Jalil},
  title   = {Fostering Safety Communication among Construction Workers: Role of Safety Climate and Crew-Level Cohesion},
  journal = {International Journal of Environmental Research and Public Health},
  year    = {2018},
  volume  = {16},
  pages   = {71},
  doi     = {10.3390/IJERPH16010071}
}

@article{qing2025,
  author  = {Qing, Siyu and Valliant, Richard},
  title   = {Extending Cochran's Sample Size Rule to Stratified Simple Random Sampling with Applications to Audit Sampling},
  journal = {Journal of Official Statistics},
  year    = {2025},
  volume  = {41},
  pages   = {309--328},
  doi     = {10.1177/0282423X241277054}
}

@inproceedings{ramponi2020,
  author    = {Ramponi, Alan and Plank, Barbara},
  title     = {Neural Unsupervised Domain Adaptation in {NLP} -- A Survey},
  booktitle = {Proceedings of the 28th International Conference on Computational Linguistics (COLING)},
  year      = {2020},
  pages     = {6838--6855},
  doi       = {10.18653/V1/2020.COLING-MAIN.603}
}

@book{rosenberg1960,
  author    = {Rosenberg, Milton J. and Hovland, Carl I.},
  title     = {Attitude Organization and Change: An Analysis of Consistency among Attitude Components},
  publisher = {Yale University Press},
  address   = {New Haven, CT},
  year      = {1960}
}

@article{rosenstock1974,
  author  = {Rosenstock, Irwin M.},
  title   = {The Health Belief Model and Preventive Health Behavior},
  journal = {Health Education \& Behavior},
  year    = {1974},
  volume  = {2},
  pages   = {354--386},
  doi     = {10.1177/109019817400200405}
}

@article{sammour2026,
  author  = {Sammour, Farouq and Zhang, Yuxin and Wang, Xin and Hu, Ming and Zhang, Zhenyu},
  title   = {Operationalizing Expert Knowledge in Construction Safety Training Through a Constraint-Guided Multi-Agent {LLM} System},
  journal = {SSRN Working Paper},
  year    = {2026},
  doi     = {10.2139/SSRN.6552781}
}

@article{schwartz1992,
  author  = {Schwartz, Shalom H.},
  title   = {Universals in the Content and Structure of Values: Theoretical Advances and Empirical Tests in 20 Countries},
  journal = {Advances in Experimental Social Psychology},
  year    = {1992},
  volume  = {25},
  pages   = {1--65},
  doi     = {10.1016/S0065-2601(08)60281-6}
}

@article{shin2014,
  author  = {Shin, Minki and Lee, Hyun-Soo and Park, Moonseo and Moon, Myunggi and Han, Sangwon},
  title   = {A system dynamics approach for modeling construction workers' safety attitudes and behaviors},
  journal = {Accident Analysis \& Prevention},
  year    = {2014},
  volume  = {68},
  pages   = {95--105},
  doi     = {10.1016/J.AAP.2013.09.019}
}

@article{sim2005,
  author  = {Sim, Julius and Wright, Chris C.},
  title   = {The Kappa Statistic in Reliability Studies: Use, Interpretation, and Sample Size Requirements},
  journal = {Physical Therapy},
  year    = {2005},
  volume  = {85},
  pages   = {257--268},
  doi     = {10.1093/PTJ/85.3.257}
}

@article{siu2003,
  author  = {Siu, Oi-Ling and Phillips, David R. and Leung, Tat-Wing},
  title   = {Age differences in safety attitudes and safety performance in Hong Kong construction workers},
  journal = {Journal of Safety Research},
  year    = {2003},
  volume  = {34},
  pages   = {199--205},
  doi     = {10.1016/S0022-4375(02)00072-5}
}

@article{smetana2024,
  author  = {Smetana, Marek and de Salles, Lucas S. and Sukharev, Igor and Khazanovich, Lev},
  title   = {Highway Construction Safety Analysis Using Large Language Models},
  journal = {Applied Sciences},
  year    = {2024},
  volume  = {14},
  pages   = {1352},
  doi     = {10.3390/APP14041352}
}

@article{sokolova2009,
  author  = {Sokolova, Marina and Lapalme, Guy},
  title   = {A systematic analysis of performance measures for classification tasks},
  journal = {Information Processing \& Management},
  year    = {2009},
  volume  = {45},
  pages   = {427--437},
  doi     = {10.1016/J.IPM.2009.03.002}
}

@article{stergioukita2015,
  author  = {Stergiou-Kita, Mary and Mansfield, Elizabeth and Bezo, Randy and Colantonio, Angela and Garritano, Enzo and Lafrance, Marc and Lewko, John and Mantis, Steve and Moody, Joel and Power, Nicole and Theberge, Nancy and Westwood, Emile and Travers, Krista},
  title   = {Danger zone: Men, masculinity and occupational health and safety in high risk occupations},
  journal = {Safety Science},
  year    = {2015},
  volume  = {80},
  pages   = {213--220},
  doi     = {10.1016/J.SSCI.2015.07.029}
}

@article{swuste2012,
  author  = {Swuste, Paul and Frijters, Adri and Guldenmund, Frank},
  title   = {Is it possible to influence safety in the building sector? A literature review extending from 1980 until the present},
  journal = {Safety Science},
  year    = {2012},
  volume  = {50},
  pages   = {1333--1343},
  doi     = {10.1016/J.SSCI.2011.12.036}
}

@article{tajfel2004,
  author  = {Tajfel, Henri and Turner, John C.},
  title   = {The Social Identity Theory of Intergroup Behavior},
  journal = {Political Psychology},
  year    = {2004},
  pages   = {276--293},
  doi     = {10.4324/9780203505984-16}
}

@article{tam2011,
  author  = {Tam, Vivian W. Y. and Fung, Ivan W. H.},
  title   = {Behavior, Attitude, and Perception toward Safety Culture from Mandatory Safety Training Course},
  journal = {Journal of Professional Issues in Engineering Education and Practice},
  year    = {2011},
  volume  = {138},
  pages   = {207--213},
  doi     = {10.1061/(ASCE)EI.1943-5541.0000104}
}

@article{tan2004,
  author  = {Tan, Pang-Ning and Kumar, Vipin and Srivastava, Jaideep},
  title   = {Selecting the right objective measure for association analysis},
  journal = {Information Systems},
  year    = {2004},
  volume  = {29},
  pages   = {293--313},
  doi     = {10.1016/S0306-4379(03)00072-3}
}

@article{thapa2025,
  author  = {Thapa, Surendrabikram and Shiwakoti, Surabhi and Shah, Sristi Bashyal and Adhikari, Subash and Veeramani, Hariram and Nasim, Mehwish and Naseem, Usman},
  title   = {Large language models ({LLM}) in computational social science: prospects, current state, and challenges},
  journal = {Social Network Analysis and Mining},
  year    = {2025},
  volume  = {15},
  pages   = {4},
  doi     = {10.1007/S13278-025-01428-9}
}

@misc{vivas2024,
  author = {{Vivas.AI}},
  title  = {Mastering prompt engineering: A guide to the {CO-STAR} and {TIDD-EC} frameworks},
  year   = {2024},
  howpublished = {\url{https://tinyurl.com/ekvhsjf6}}
}

@article{wang2026,
  author  = {Wang, Yang and Liu, Chao and Zhao, Wei},
  title   = {Safety attitudes and safety management: perspectives of managerial personnel and frontline workers in construction sites in China},
  journal = {Engineering, Construction and Architectural Management},
  year    = {2026},
  volume  = {33},
  pages   = {666--689},
  doi     = {10.1108/ECAM-08-2024-1032}
}

@misc{wei2025,
  author = {Wei, Dong and Mao, Minghui and Fang, Xin and Chau, Michael},
  title  = {Short-{PHD}: Detecting Short LLM-generated Text with Topological Data Analysis After Off-topic Content Insertion},
  year   = {2025},
  note   = {arXiv preprint}
}

@article{wilson2000,
  author  = {Wilson, Timothy D. and Lindsey, Samuel and Schooler, Tonya Y.},
  title   = {A model of dual attitudes},
  journal = {Psychological Review},
  year    = {2000},
  volume  = {107},
  pages   = {101--126},
  doi     = {10.1037/0033-295X.107.1.101}
}

@article{xu2023,
  author  = {Xu, Sheng and Lin, Boyu and Zou, Patrick X. W.},
  title   = {Examining construction group's safety attitude resilience under major disruptions: An agent-based modelling approach},
  journal = {Safety Science},
  year    = {2023},
  volume  = {161},
  pages   = {106071},
  doi     = {10.1016/J.SSCI.2023.106071}
}

@article{xu2018,
  author  = {Xu, Sheng and Zou, Patrick X. W. and Luo, Hanbin},
  title   = {Impact of Attitudinal Ambivalence on Safety Behaviour in Construction},
  journal = {Advances in Civil Engineering},
  year    = {2018},
  volume  = {2018},
  pages   = {7138930},
  doi     = {10.1155/2018/7138930}
}

@article{yao2021,
  author  = {Yao, Qiangqiang and Li, Rita Yi Man and Song, Li and Crabbe, M. James C.},
  title   = {Construction safety knowledge sharing on Twitter: A social network analysis},
  journal = {Safety Science},
  year    = {2021},
  volume  = {143},
  pages   = {105411},
  doi     = {10.1016/J.SSCI.2021.105411}
}

@article{yuan2022,
  author  = {Yuan, Bo and Xu, Sheng and Chen, Lei and Niu, Miaomiao},
  title   = {How Do Psychological Cognition and Institutional Environment Affect the Unsafe Behavior of Construction Workers? Research on fsQCA Method},
  journal = {Frontiers in Psychology},
  year    = {2022},
  volume  = {13},
  pages   = {875348},
  doi     = {10.3389/FPSYG.2022.875348}
}

@article{zhou2022,
  author  = {Zhou, Mi and Chen, Xi and He, Li and Ouedraogo, Frank A. K.},
  title   = {Dual-Attitude Decision-Making Processes of Construction Worker Safety Behaviors: A Simulation-Based Approach},
  journal = {International Journal of Environmental Research and Public Health},
  year    = {2022},
  volume  = {19},
  pages   = {14413},
  doi     = {10.3390/IJERPH192114413}
}

\end{document}